\documentclass{article}

\PassOptionsToPackage{numbers, compress}{natbib}

\usepackage{iclr2025_conference,times}

\usepackage[utf8]{inputenc} 
\usepackage[T1]{fontenc}    
\usepackage{url}            
\usepackage{booktabs}       
\usepackage{amsfonts}       
\usepackage{nicefrac}       
\usepackage{microtype}      
\usepackage{xcolor}         

\newcommand{\etal}{\textit{et al.}\xspace} 

\usepackage{graphicx}
\usepackage{booktabs}
\usepackage{kotex}
\usepackage{bm}

\usepackage{wrapfig}
\usepackage{multirow}
\usepackage[pagebackref,breaklinks,colorlinks]{hyperref}
\usepackage{booktabs}
\usepackage{colortbl}
\usepackage{tabulary}
\usepackage{algorithmicx}
\usepackage{algpseudocode}
\usepackage{listings}
\usepackage{multicol}
\usepackage{xspace}
\usepackage{subcaption}
\usepackage{graphbox}
\usepackage{arydshln}
\usepackage{caption}
\usepackage{algpseudocode}
\usepackage{algorithm}
\usepackage{adjustbox}
\usepackage{amsmath}
\usepackage{microtype}

\usepackage{diagbox}

\iclrfinalcopy

\title{Denoising Task Difficulty-based Curriculum for Training Diffusion Models}

\author{
  Jin-Young Kim\thanks{Co-first author \quad $^\dagger$Corresponding author}\;$^\dagger$ \quad Hyojun Go$^{*}$ \quad Soonwoo Kwon$^{*}$ \quad Hyun-Gyoon kim$^{1\dagger}$
  \vspace{0.3mm}
  \\ Ajou University$^1$
  \\ \tt\small \{seago0828, gohyojun15, swkwon.john\}@gmail.com, hyungyoonkim@ajou.ac.kr
}

\begin{document}
\addtocontents{toc}{\protect\setcounter{tocdepth}{0}}

\maketitle
\newcommand{\todoc}[2]{{\textcolor{#1}{\textbf{#2}}}}
\newcommand{\todoblue}[1]{\todoc{blue}{\textbf{#1}}}
\newcommand{\todored}[1]{\todoc{red}{\textbf{#1}}}

\newcommand{\todogreen}[1]{\todoc{green}{\textbf{#1}}}

\newcommand{\hyojun}[1]{\todoblue{\textbf{hyojun:} #1}}
\newcommand{\jy}[1]{\todored{\textbf{jeremy:} #1}}
\newcommand{\snoopy}[1]{\todogreen{soonwoo: #1}}

\begin{abstract}

Diffusion-based generative models have emerged as powerful tools in the realm of generative modeling. 
Despite extensive research on denoising across various timesteps and noise levels, a conflict persists regarding the relative difficulties of the denoising tasks. 
While various studies argue that lower timesteps present more challenging tasks, others contend that higher timesteps are more difficult. 
To address this conflict, our study undertakes a comprehensive examination of task difficulties, focusing on convergence behavior and changes in relative entropy between consecutive probability distributions across timesteps.
Our observational study reveals that denoising at earlier timesteps poses challenges characterized by slower convergence and higher relative entropy, indicating increased task difficulty at these lower timesteps.
Building on these observations, we introduce an easy-to-hard learning scheme, drawing from curriculum learning, to enhance the training process of diffusion models. 
By organizing timesteps or noise levels into clusters and training models with ascending orders of difficulty, we facilitate an order-aware training regime, progressing from easier to harder denoising tasks, thereby deviating from the conventional approach of training diffusion models simultaneously across all timesteps.
Our approach leads to improved performance and faster convergence by leveraging benefits of curriculum learning, while maintaining orthogonality with existing improvements in diffusion training techniques.
We validate these advantages through comprehensive experiments in image generation tasks, including unconditional, class-conditional, and text-to-image generation.
\end{abstract}
\section{Introduction}

Diffusion-based generative models~\citep{ho2020denoising,sohl2015deep,song2021scorebased} have achieved significant advancements in the realm of generative tasks, demonstrating notable success across various fields such as image~\citep{dhariwal2021diffusion}, video~\citep{ho2022imagen,harvey2022flexible}, and 3D~\citep{woo2023harmonyview,liu2023syncdreamer} generation.
Specifically, their exceptional adaptability and promising performance in diverse image generation contexts, such as unconditional~\citep{karras2022elucidating, nichol2021improved}, class-conditional~\citep{dhariwal2021diffusion}, and text-conditional scenarios~\citep{balaji2022ediffi, ramesh2022hierarchical}, demonstrate their significant impact.
Such achievements have led to a growing interest in further deepening the analysis and enhancing diffusion models.

Diffusion models~\citep{ho2020denoising, song2021scorebased} are designed to reverse the corruption of the data through the learning process at different noise levels and over multiple timesteps.
Recent works have delved into the learning of diffusion models across various noise levels and timesteps, revealing different stages of diffusion models.
For example, Choi~\etal~\citep{choi2022perception} observe that when a diffusion model performs a denoising task from large to small timestep, it first generates coarse features, then gradually generates perceptually rich content, and later refines the details.
Similar observation is also identified in text-to-image diffusion models~\citep{balaji2022ediffi}.
Besides this aspect, various studies have further explored the learning of diffusion models across timesteps and noise levels, elucidating their transition from denoising to generative functionalities~\citep{deja2022analyzing}, modular attributes~\citep{yue2024exploring}, frequency characteristics~\citep{yang2023diffusion, lee2023multi}, trajectories~\citep{pan2023t}, affinity~\citep{go2023addressing}, and variations of targets~\citep{xu2023stable}.

These observations have not only deepened understanding of diffusion models but have also directly contributed to improvement in diffusion models.
Specifically, these insights are incorporated into their method design in various works, including loss functions~\citep{hang2023efficient,xu2023stable}, architectures~\citep{lee2023multi,balaji2022ediffi}, accelerated sampling~\citep{pan2023t}, representations~\citep{yue2024exploring}, and guidance~\citep{go2023towards}.
Given the tangible benefits already realized from such studies, further in-depth analysis of diffusion models across timesteps and noise levels is crucial for uncovering insights and achieving unprecedented advancements in their capabilities.

In this paper, to enrich the current understanding across various timesteps and noise levels, we investigate under-explored areas within diffusion models focusing on the \textit{task difficulties} of denoising.
Regarding denoising task difficulties, previous works speculate that denoising tasks across timesteps have different difficulties~\citep{li2023autodiffusion,balaji2022ediffi}, yet a detailed exploration of these variances remains sparse.
Moreover, there exists a notable discrepancy among studies, with works identifying larger timesteps as more difficult~\citep{ho2020denoising,hang2023efficient}, while others argue that smaller timesteps pose greater difficulties~\citep{karras2022elucidating, dockhorn2021score, kim2022soft}.
The discrepancy in difficulty across timesteps not only impedes the accurate interpretation of previous studies but also hinders the development of sophisticated training methods that properly utilize the timestep-wise variation in difficulty.

In this regard, we first analyze task difficulties in two aspects to resolve these conflicts: 1) convergence properties in the learning of denoising tasks at each timestep, and 2) the change in relative entropy between consecutive probability distributions over timesteps.
In the first aspect, our analysis reveals distinct convergence behaviors across timesteps, demonstrating that models trained on larger timesteps exhibit faster convergence.
In the second aspect, we also observe a decrease in relative entropy as we progress to later timesteps.
By integrating these, we confirm that denoising tasks at earlier timesteps are more difficult, indicated by slower convergence speeds and greater changes in relative entropy.

Furthermore, building on these observations, we integrate an easy-to-hard learning scheme, a concept well-established in the curriculum learning literature~\citep{hacohen2019power, kong2021adaptive, chang-etal-2021-order, wang-etal-2020-curriculum, pentina2015curriculum}, into the training process of diffusion models.
Specifically, we organize timesteps or noise levels into clusters and train the diffusion models with ascending levels of difficulty, moving from clusters categorized by higher to lower timesteps.
After this curriculum process, models simultaneously learn whole timesteps as standard diffusion training~\citep{ho2020denoising, song2021scorebased, ho2022classifier} to reach the convergence point. 
Unlike conventional approaches where diffusion models are trained simultaneously across all timesteps, our method distinguishes itself by incorporating a sequential, order-aware training regime, reflecting an intended progression from easier to harder denoising tasks.

Building upon this foundation, our curricular approach offers several notable advantages: \textbf{1) Improved Performance} and \textbf{2) Faster Convergence:} By leveraging the inherent benefits of curriculum learning, our method significantly enhances the quality of generation and the speed of convergence. 
\textbf{3) Orthogonality with Existing Improvements:} Our approach is inherently model-agnostic, ensuring broad applicability across various diffusion models. Additionally, it can be integrated with advanced diffusion training techniques, such as loss weighting~\citep{choi2022perception, hang2023efficient, go2023addressing, karras2023analyzing}.

Finally, we empirically validate the advantages of our method by conducting comprehensive experiments across a variety of image-generation tasks. These include unconditional generation, class-conditional generation, and text-to-image generation, utilizing datasets such as FFHQ~\citep{karras2019style}, ImageNet~\citep{deng2009imagenet}, and MS-COCO~\citep{lin2014microsoft}. 
By integrating our curriculum learning strategy into architectures— DiT~\citep{peebles2022scalable}, EDM~\citep{karras2022elucidating}, and SiT~\citep{ma2024sit}—we demonstrate the efficacy of our approach in enhancing performance, accelerating convergence speed, and maintaining compatibility with existing techniques.



\vspace{-0.1cm}
\section{Related Works}
\vspace{-0.2cm}

\subsection{Diffusion Models}
\vspace{-0.2cm}
Diffusion models~\citep{ho2020denoising,sohl2015deep,song2021scorebased} are a group of generative models that create samples by utilizing a learned denoising process to noise.
Several works have focused on improving diffusion models in various aspects, including model architectures~\citep{park2024denoising,dhariwal2021diffusion, park2024switch}, sampling speed~\citep{song2020denoising,lu2022dpm,liu2023oms}, training objectives~\citep{hang2023efficient, choi2022perception, go2023addressing, kingma2024understanding, ma2024sit}. 
These endeavors often involve investigating what diffusion models learn by dividing its process, aiming to enhance the performance of diffusion models. 
P2~\citep{choi2022perception} under-weights training loss functions at the clean-up stage from their observation that diffusion models learn coarse, perceptual, and removing noises at large, medium, and small timesteps.
Ediff-I~\citep{balaji2022ediffi} observes that earlier sampling parts rely on conditions for generation, whereas later parts ignore the conditions. They employ multiple denoisers to address the diverse characteristics of tasks associated with different parts of the sampling process.
Moreover, various works have also investigated these aspects related to timesteps~\citep{deja2022analyzing, yue2024exploring, yang2023diffusion, lee2023multi, pan2023t, go2023addressing, xu2023stable} (detailed illustrations can be found in Appendix \textcolor{red}{A}). 
While our study aligns with the above works, we analyze the under-explored aspect of denoising task difficulty.
Furthermore, we leverage these observations to propose a curriculum learning approach.
\vspace{-0.2cm}

\subsection{Denoising Difficulties on Diffusion Models}
\vspace{-0.1cm}
Difficulties in denoising tasks in diffusion have been referred to by various works, but this aspect is not deeply explored.
Several studies hypothesize that denoising tasks in diffusion encompass diverse difficulties~\citep{li2023autodiffusion,balaji2022ediffi}, and there have been conflicts regarding these difficulties between previous works.

Certain studies consider denoising at larger noise levels and timesteps to be more difficult~\citep{ho2020denoising,hang2023efficient}, the focus is on the challenges associated with reconstructing data from substantial noise. 
For instance, Hang~\etal~\citep{hang2023efficient} articulate that while smaller timesteps (approaching zero) may require straightforward reconstructions, such strategies become less effective at higher noise levels or in larger timesteps. 
Similarly, Ho~\etal~\citep{ho2020denoising} elucidate that their approach de-emphasizes loss terms at smaller timesteps to prioritize learning on the more challenging tasks at larger timesteps, thereby enhancing sample quality.
Conversely, other studies argue that earlier timesteps or lower noise levels also present significant challenges. 
Karras~\etal\cite{karras2022elucidating} suggest that detecting noise at low levels is challenging due to its minimal presence.
Also, Kim~\etal~\citep{kim2022soft} illustrate the increasing difficulty and high variance in score estimation as timesteps approach zero, disturbing stable training of models.
In line with these observations, Dockhorn~\etal~\citep{dockhorn2021score} build upon insights of~\citep{kim2022soft}, acknowledging the complexities at near zero timesteps, where score becomes highly complex and potentially unbounded. 

In this work, we aim to resolve this conflict through an in-depth analysis of convergence properties and changes in relative entropy between consecutive probability distributions across timesteps.

\vspace{-0.2cm}
\subsection{Curriculum Learning}
\vspace{-0.2cm}
Curriculum learning~\citep{bengio2009curriculum, hacohen2019power, kong2021adaptive}, inspired by human learning patterns, is a method of training models in a structured order, starting with easier tasks~\citep{pentina2015curriculum} or examples~\citep{bengio2009curriculum} and gradually increasing difficulty. 
As pointed out by~\citep{bengio2009curriculum}, curriculum learning formulation can be viewed as a continuation method~\citep{allgower2003introduction}, which starts from a smoother objective and gradually transformed into a less smooth version until it reaches the original objective function.
Through this foundation, various works have achieved improved performance and faster convergence compared to standard training based on random mini-batches sampled uniformly~\citep{hacohen2019power, kong2021adaptive, chang-etal-2021-order, wang-etal-2020-curriculum}.

Curriculum learning primarily comprises two components: a curriculum scoring function, measuring the difficulty of tasks or examples, and a pacing function, modulating the speed of the curriculum progress. 
Regarding a curriculum score function, early studies have utilized human intuition for measuring difficulty, such as the complexity of geometric shapes in images~\citep{bengio2009curriculum} or the length of sequences~\citep{spitkovsky2010baby}.
Recently, various works employ models to measure difficulty, including confidence of pre-trained models~\citep{hacohen2019power} and the loss of the current models~\citep{kong2021adaptive}.
For the pacing function, a predefined pacing function has been employed, which involves training using a predetermined curriculum progression~\citep{hacohen2019power, wu2020curricula}. There are various forms of this and they can be generally represented as a function of training iteration~\citep{hacohen2019power, wu2020curricula}.
Contrary to this, there have been proposals for pacing techniques dynamically adjusting based on the loss or performance of the current model during training~\citep{kumar2010self,jiang2014self}.

In the diffusion model literature, curriculum learning has been utilized to organize the order of training data types based on prior knowledge of targeted generation tasks~\citep{tang2023anytoany, yang2023diffsound}. 
Tang~\etal~\citep{tang2023anytoany} sequentially train video diffusion models with lower resolution and FPS datasets before progressing to higher resolution and FPS datasets. 
Similarly, Yang~\etal~\citep{yang2023diffsound} order text-to-sound generation data based on the number of events in audio clips, training diffusion models from lower to higher events datasets. 
In contrast, our method explores the nature of denoising task difficulty in diffusion models and proposes a curriculum learning approach that progresses from easy to hard timesteps, deviating from the standard simultaneous training of all timesteps.
Also, while consistency models~\citep{song2023consistency, song2023improved} adopt a curriculum approach to discretizing noise levels, progressively increasing the discretization steps of noise levels during training, we have distinct by exploring which noise level should be learned first and investigating the difficulty of denoising at each noise level.

\vspace{-0.1cm}
\section{Preliminaries}
\vspace{-0.2cm}
In this section, we provide the necessary background on diffusion models~\citep{ho2020denoising,sohl2015deep,song2021scorebased}.
Let $\bm{x}_0 \in \mathbb{R}^d$ be a sample from the data distribution $p_0(\bm{x})$.
The forward process of diffusion models transforms data $\bm{x}_0$ to latent $\bm{x}_{t \in [0, T]}$ by iteratively adding Gaussian noise. 
This can be formulated as a stochastic differential equation (SDE)~\citep{song2021scorebased} as $\text{d}\bm{x}_t = f(t) \bm{x}_t \text{d}t  + g(t) \text{d} \bm{w}_t$, 
where $f(t)$ and $g(t)$ are drift and diffusion coefficients, and $\bm{w}_t$ is the standard Wiener process.
The Gaussian transition kernel of this SDE is formulated as:
\begin{equation}
    p_{0t}(\bm{x}_t|\bm{x}_0) = \mathcal{N}(\bm{x}_t; s_t \bm{x}_0, s_t^2\sigma_t^2 \mathbf{I}), \;\;  s_t = \text{exp} \left(\int_0^t f(\xi)\text{d}\xi \right), \;\;  \sigma_t = \sqrt{\int_0^t\frac{g(\xi)^2}{s_\xi^2} \text{d}\xi}.
\end{equation}
For generation, diffusion models aim to learn the corresponding reverse SDE represented as:
\begin{equation}
    \text{d}\bm{x}_t = \left[  f(t)\bm{x}_t - g^2(t) \nabla \log p_t(\bm{x}_t)\right]\text{d}\bar{t} +g(t)\text{d}\bar{\bm{w}}_t,
\end{equation}
where $\bar{\bm{w}}_t$ and $\text{d}\bar{t}$ denote the reverse-time Wiener process and the infinitesimal reverse-time, respectively, with the actual data score $\nabla \log p_t(\bm{x}_t)$.
In most cases, a neural network $\epsilon_\theta$ having parameter $\theta$ is utilized to approximate this score function by learning the denoising tasks for each timestep $t$ from score matching loss $\mathcal{L}$~\citep{song2019generative}:
\begin{equation}
\label{eq:score_matching_loss}
    \mathcal{L} = \frac{1}{2}\int^T_0 \mathcal{L}_t  \text{d}t, \quad \mathcal{L}_t = \omega(t) \mathbb{E}_{\bm{x}_t \sim p_{0t} (\bm{x}_t | \bm{x}_0), \bm{x_0} \sim p_0} \left[ ||\epsilon_\theta(\bm{x}_t, t) - \nabla \log p_{0t}(\bm{x}_t| \bm{x_}0)||_2^2 \right],
\end{equation}
where $\omega(t)$ is loss weights for $t$ and $p_{0t}(\bm{x}_t | \bm{x}_0)$ is the transition density of $\bm{x}_t$ from the initial timestep $0$ to $t$.
This object can be interpreted as a noise-matching loss in DDPM~\citep{ho2020denoising}, which predicts noise components in $\bm{x}_t$ and can be illustrated as $\int_0^T\mathbb{E}_{\bm{x}_0\sim p_0, \epsilon \sim \mathcal{N}(0, \mathbf{I})}[ ||\epsilon_\theta (\sqrt{\bar{\alpha}_t} \bm{x}_0 + \sqrt{1-\bar{\alpha}_t} \epsilon, t) - \epsilon||_2^2 ] \text{d}t$.
This is regularly denoted as $\epsilon$-prediction parameterization~\citep{ho2020denoising,jabri2022scalable}, and several other parameterizations including $F$-prediction~\citep{karras2022elucidating, kingma2024understanding}, score-prediction~\citep{song2021scorebased} and velocity-prediction~\citep{ma2024sit} have been proposed.

\vspace{-0.1cm}
\section{Observations}
\vspace{-0.2cm}
\label{sec:observations}

In this section, we examine the difficulties associated with learning denoising tasks across different timesteps, addressing inconsistencies in prior works regarding these difficulties.
Our analysis is structured around two key aspects: 1) the convergence of loss and denoising performance across timesteps, providing insights into learning dynamics at various timestep stages in Section~\ref{sec:analysis_task_difficulty_convergence}; and 2) the relative entropy change from $p_t$ to $p_{t-1}$ as a function of $t$, offering a quantitative measure of task difficulty progression over $t$ in Section~\ref{sec:analysis_relative_entropy}. 
Upon integrating our findings, we establish a key conclusion: the learning difficulty for denoising tasks escalates as the timestep $t$ decreases.

\subsection{Analysis on the Task Difficulty in terms of Convergence Speed}
\label{sec:analysis_task_difficulty_convergence}


In this study, we analyze the convergence speed of loss and denoising performance across timesteps. 
To comprehensively cover various diffusion parameterizations, we utilized the notable frameworks DiT~\citep{jabri2022scalable} for $\epsilon$-prediction, EDM~\citep{karras2022elucidating} for $F$-prediction, SiT~\citep{ma2024sit} for velocity prediction.
Detailed descriptions of the experimental setups are provided in Appendix~\textcolor{red}{B}.

\begin{figure}[t]
    \centering
    \begin{subfigure}[b]{0.8\textwidth}
         \centering
         \includegraphics[width=\textwidth]{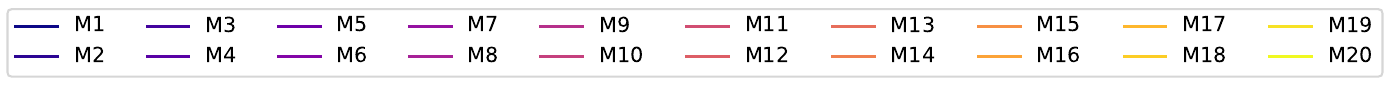}
         \vspace{-4.5mm}
    \end{subfigure}
    \begin{subfigure}[b]{0.32\textwidth}
         \centering
         \includegraphics[width=\textwidth]{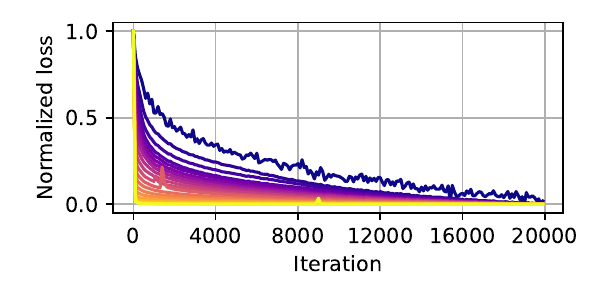}
         \vspace{-7mm}
         \caption{DiT (Loss convergence)}
         \label{fig:loss_conv_dit}
    \end{subfigure}
    \hspace{-3mm}
    \begin{subfigure}[b]{0.32\textwidth}
         \centering
         \vspace{-0.5mm}
         \includegraphics[width=\textwidth]{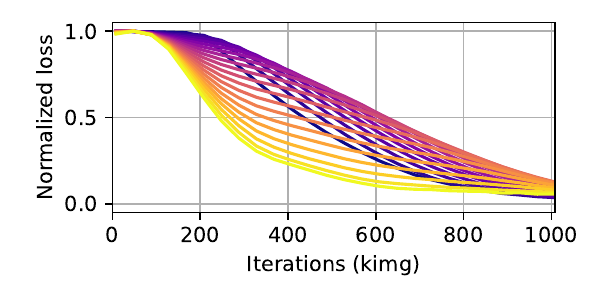}
         \vspace{-7mm}
         \caption{EDM (Loss convergence)}
         \label{fig:loss_conv_edm}
    \end{subfigure}
    \hspace{-3mm}
    \begin{subfigure}[b]{0.31\textwidth}
         \centering
         \vspace{-9mm}
         \includegraphics[width=\textwidth]{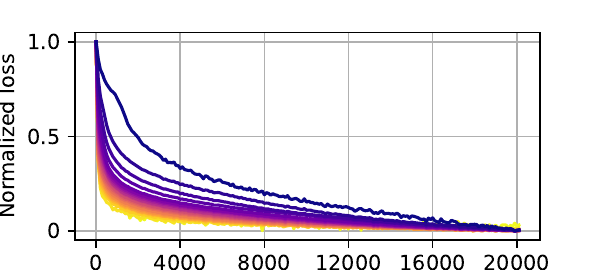}
         \vspace{-4mm}
         \caption{SiT (Loss convergence)}
         \label{fig:loss_conv_sit}
    \end{subfigure}
    \hspace{-7mm}
    \begin{subfigure}[b]{0.32\textwidth}
         \centering
         \includegraphics[width=\textwidth]{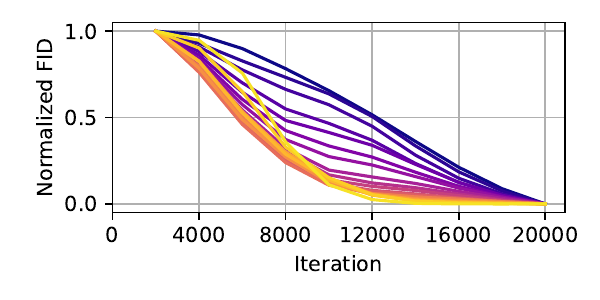}
         \vspace{-7mm}
         \caption{DiT (Task convergence)}
         \label{fig:fid_conv_dit}
    \end{subfigure}
    \hspace{-3mm}
    \begin{subfigure}[b]{0.31\textwidth}
         \centering
         \vspace{-0.5mm}
         \includegraphics[width=\textwidth]{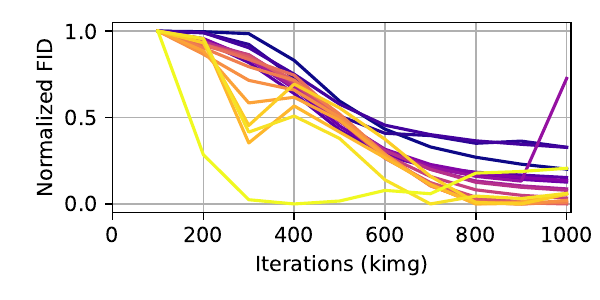}
         \vspace{-7mm}
         \caption{EDM (Task convergence)}
         \label{fig:fid_conv_edm}
    \end{subfigure}
    \hspace{-2mm}
    \begin{subfigure}[b]{0.31\textwidth}
         \centering
         \includegraphics[width=\textwidth]{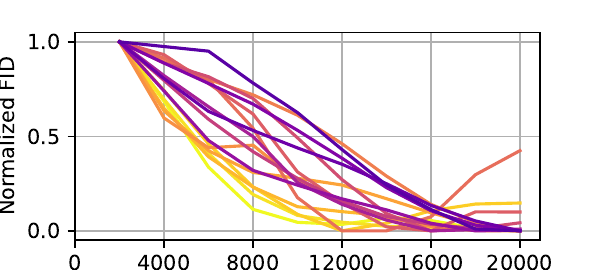}
         \vspace{-4mm}
         \caption{SiT (Task convergence)}
         \label{fig:fid_conv_sit}
    \end{subfigure}
    \vspace{-0.4cm}
    \caption{
    Loss and FID convergence plotted during training for each diffusion model $\mathrm{M}_i$ in DiT, EDM, and SiT. Since the loss scale for each model is different, we show the normalized value. We observe that as $i$ increases (i.e., corresponding to larger denoising timesteps), the loss converges more rapidly, and this convergence speed correlates with that of the FID scores. 
    }
    \vspace{-0.4cm}
    \label{fig:convergence}
\end{figure}

\paragraph{Convegence speed on loss.} 
First, we analyze convergence characteristics of training loss across timesteps $t$. 
We divided whole timesteps $[0, T]$ into 20 uniformly divided intervals and trained 20 models $\{\mathrm{M}_i\}_{i=1}^{20}$ where $i$-th model learns denoising tasks in $[\frac{i-1}{20}T, \frac{i}{20}T]$ for DiT and SiT, $[\mathrm{\Phi}^{-1}(\frac{i-1}{N}), \mathrm{\Phi}^{-1}(\frac{i}{N})]$ for EDM where $\mathrm{\Phi}^{-1}$ is the inverse cumulative distribution function of the Gaussian distribution.
During training, we tracked the loss values through iterations and plotted their convergence speed by normalizing their value in Figs.~\ref{fig:loss_conv_dit}-\ref{fig:loss_conv_sit}.
As shown in the results, it is apparent that as $i$ increases towards $i=20$, the convergence accelerates in both DiT, EDM, and SiT, suggesting that models learning larger timesteps can reach convergence more swiftly and reinforcing the notion that denoising tasks with larger timesteps are less difficult. 

\paragraph{Convegence speed on denoising performance.} 
We also delve deeper into a convergence of denoising performance according to timesteps with 20 distinct models $\{\mathrm{M}_i\}_{i=1}^{20}$.
To evaluate the performance of denoising tasks of each model, we generated samples where $\mathrm{M}_i$ was employed for denoising within the timesteps that it was trained on, while a diffusion model learned whole timesteps handled denoising for the remaining timesteps as in~\citep{go2023addressing}. 
Then, the performance of the denoising capability of $\mathrm{M}_i$ can be quantitatively measured using the FID score~\citep{heusel2017gans}, enabling us to observe the performance convergence of each model on denoising tasks throughout the training process. 
Figures~\ref{fig:fid_conv_dit}-\ref{fig:fid_conv_sit} depict this convergence. 
They illustrate that, similar to loss convergence experiments, denoising performance converges more swiftly for models $\mathrm{M}_i$ with larger $i$ values, as observed across the DiT, EDM, and SiT.
These results also suggest that models trained on later timesteps, indicated by larger $i$ values, achieve faster convergence, highlighting easier task difficulty at larger timesteps.

\subsection{Exploration on Difficulties of Denoising Tasks}
\label{sec:analysis_relative_entropy}

\begin{wrapfigure}[9]{r}{0.33\textwidth}
    \centering
    \vspace{-8mm}
    \includegraphics[width=1\linewidth]{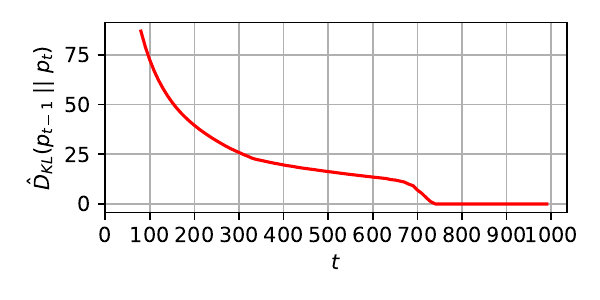}
    \vspace{-8mm}
    \caption{
    The KLD of $p_{t-1}$ from $p_t$ against denoising timestep. As the timestep increases, the dynamics decrease. 
    }
    \label{fig:dynamics}
\end{wrapfigure}
Beyond empirical convergence metrics, we also delve into analyzing the relative entropy between $p_t$ and $p_{t-1}$ to better understand task difficulties from a distributional perspective.
The training of diffusion models implicitly involves learning the distribution of the reverse process of the corresponding SDE.
To be specific, the transition probability of the reverse process is expressed as a conditional normal distribution whose mean parameter is modeled by neural networks, and they are thereby trained to learn the dynamics of the reverse process~\citep{ho2020denoising}.
Furthermore, an unconditional distribution of $\bm{x}_t$ can be obtained by marginalizing transition densities over the prior distribution, indicating that information on the dynamics of the marginal distribution is fed to neural networks~\citep{song2021scorebased}.

To analyze the relationship between the dynamics of the unconditional distribution and the rate of loss convergence, we use the Kullback-Leibler (KL) divergence of $p_{t-1}$ to $p_{t}$, $D_{KL}(p_{t-1} || p_t)$, as a quantitative measure. It is a pertinent divergence in that the training mechanism of diffusion models involves maximizing the likelihood of the reverse process.
The KL divergence $D_{KL} (p_{t-1} || p_t)$ is given by $D_{KL} \left(p_{t-1} || p_t \right) = \mathbb{E}_{\bm{x} \sim p_{t-1}} \left[\log \left(\frac{p_{t-1} (\bm{x})}{p_t (\bm{x})} \right)\right]$.
Moreover, the distribution $p_t$ of $\bm{x}$ at $t$ is expressed by $p_t (\bm{x}_t) = \int p_{0t} (\bm{x}_t | \bm{x}_0=\bm{y}) p_0 (\bm{y}) \text{d}y=\mathbb{E}_{\bm{x}_0\sim p_0} \left[ p_{0t} \left( \bm{x}_t | \bm{x}_0\right) \right]$.
However, since the explicit density form of $p_0$ is unknown and it is computationally infeasible to estimate high-dimensional integrals, we approximate them through unbiased estimators (details in Appendix~\textcolor{red}{C}).
The empirical results of $D_{KL} (p_{t-1} || p_t)$ for $64\times64$ image data are given in Fig.~\ref{fig:dynamics}.
As seen, the relative entropy tends to decrease as $t$ increases (i.e., $D_{KL} \left(p_{s-1} || p_s \right) \le D_{KL} \left(p_{t-1} || p_t \right)$ for $s\le t$), which is consistent with the results in Fig.~\ref{fig:convergence}.

This observation may stem from the inherent low-dimensional manifold of image data.
As is well-known (e.g.,~\citep{ruderman1994statistics}), the image data is distributed on a relatively low-dimensional manifold with a narrow support and a highly peaked multi-modal structure.
On the other hand, as Gaussian noise is iteratively added, the distribution of $\bm{x}_t$ approaches the independent Gaussian distribution in the ambient space. 
Consequently, the support of the manifold broadens and the score function becomes regular over the ambient space with increasing $t$. 
This nature of the unconditional distribution may cause the relative entropy from $p_t$ to $p_{t-1}$ to decrease with $t$, indicating that it is more difficult to accurately represent the dynamics of the reverse process at small $t$. More discussion is in Appendix~\textcolor{red}{C}.

\vspace{-0.1cm}
\section{Methodology}
\vspace{-0.2cm}
In Section~\ref{sec:observations}, we observe that denoising tasks at smaller $t$ are more difficult to learn by models.
From these order of difficulties in denoising tasks, we propose the incorporation strategy of an easy-to-hard training scheme, that has demonstrated its effectiveness in curriculum literature~\citep{bengio2009curriculum, hacohen2019power, kong2021adaptive}, for improving diffusion models' training.


\subsection{Design of Curriculum Learning in Diffusion Models}
\label{sec:method curriculum stages}

\begin{figure}[t!]
\centering
\includegraphics[width=0.95\linewidth]{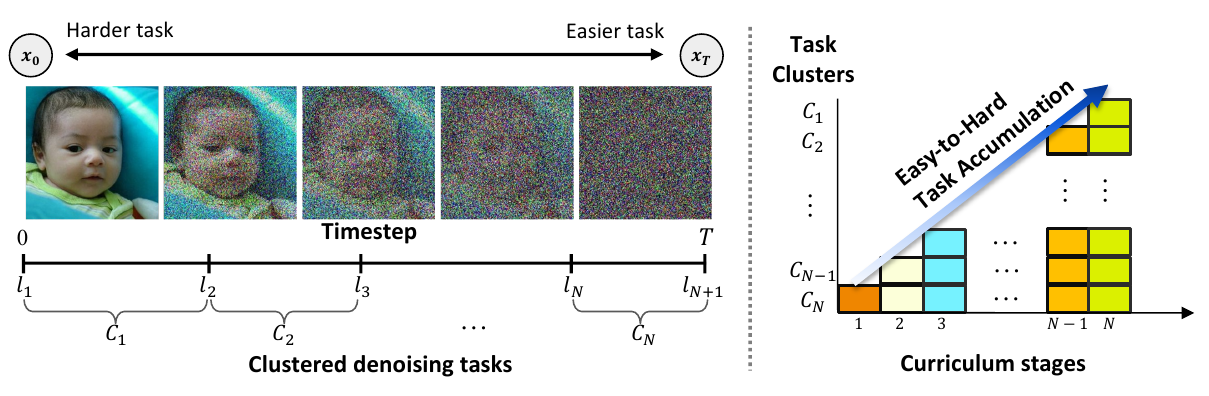}
\vspace{-1mm}
\caption{The overview of our curriculum learning approach for diffusion models. \textbf{(Left)} We divide the timesteps into $N$ clusters, ${C_1, ..., C_N}$, with the difficulty of denoising tasks increasing from $C_N$ (easiest) to $C_1$ (hardest). \textbf{(Right)} As the curriculum progresses, learning accumulates harder task clusters, gradually increasing task difficulties.}
\vspace{-5mm}
\label{fig:main_figure}
\end{figure}

As we observed in Section~\ref{sec:observations}, difficulties in denoising tasks increase as $t$ gets smaller.
To utilize an easy-to-hard curriculum learning approach, we first divide the entire range of timesteps into $N$ clusters, denoted as $\{C_i\}_{i=1}^N$, where each cluster $C_i$ spans an interval $[l_i, l_{i+1}]$, ensuring $l_i < l_{i+1}$, with $l_1 = 0$ and $l_{N+1} = T$, as shown on the left side of Fig.~\ref{fig:main_figure}.
The curriculum for training is constructed by regarding these task clusters as unit tasks, starting from the least challenging (the $N$-th cluster $C_N$) and advancing towards the most difficult (the first cluster $C_1$), through $N$ distinct stages.
Specifically, in the $n$-th curriculum stage, we jointly train the model with denoising tasks sampled from the clusters $\bigcup_{j=N-(n-1)}^{N}C_j$ as illustrated in the right side of Fig.~\ref{fig:main_figure}. 
The transition of the curriculum stages is determined by the pacing function, which will be discussed in the next section.
After completing these $N$-stages of curriculum learning, the model continues to learn across the entire range of timesteps, $\bigcup_{j=1}^{N}C_j$, same as standard diffusion training.

The next consideration involves determining the boundaries for each cluster $l_i$.
A straightforward approach is to uniformly divide the entire timestep interval $[0, T]$ as $C_i = [\frac{(i-1) \cdot T}{N}, \frac{i \cdot T}{N}]$ for $i = 1, 2, \ldots, N$. 
However, this method does not account for variations in noise levels across different timesteps.
Therefore, to address this issue more effectively, we adopt an SNR-based interval clustering technique as used in~\citep{go2023addressing}, which aligns the clustering with the actual changes in noise levels, potentially enhancing curriculum learning adaptability to varying noise conditions.

For EDM~\citep{karras2019style} which operates based on the noise level $\sigma$ rather than the timestep $t$, and where $\sigma$ is sampled from a log-normal distribution such that $\log(\sigma) \sim \mathcal{N}(P_{\text{mean}}, P_{\text{std}}^2)$ during training, our clustering strategy for timesteps cannot be directly transposed.
Given the log-normal distribution of $\sigma$, dividing it directly is impractical because $\sigma$ can extend over a wide range of values. 
Instead, we adapt our clustering approach to suit the log-normal characteristics by defining noise level clusters $C_i$. 
Specifically, we delineate $C_i = [\mathrm{\Phi}^{-1}(\frac{i-1}{N}), \mathrm{\Phi}^{-1}(\frac{i}{N})]$, where $\mathrm{\Phi}^{-1}$ is the inverse cumulative distribution function (quantile function) of the Gaussian distribution $\mathcal{N}(P_{\text{mean}}, P_{\text{std}}^2)$.
This method segments the noise levels into intervals by reflecting their probabilistic distribution.

\newcommand{\mycomment}[1]{{\color{gray}\fontsize{6pt}{6pt}\selectfont\texttt{#1}}}



\subsection{Pacing Strategy of Curriculum}
\label{sec:method pacing starategy}

To effectively train the diffusion model according to the provided curriculum design, it is crucial to define a suitable \textit{pacing function} for determining the transition of each $N$ distinct curriculum. 
Training for a fixed number of iterations for each curriculum stage is the simplest implementation (We also contain this method in experiments as \textit{`NaiveCL'} in Section~\ref{sec:exp:comparative_eval}).
However, the convergence rate of each curriculum phase varies significantly, as demonstrated in Fig.~\ref{fig:convergence}. 
Hence, we propose adopting an adaptive number of iterations for each curriculum, akin to the varied exponential pacing approach explored by Hacohen~\etal~\citep{hacohen2019power}.
Our pacing function utilizes the training loss to determine transition moments and transitions to the next stage occur when the training loss converges at the current stage.
Specifically, we introduce the maximum patience iteration $\tau$, and if the loss does not improve consecutively for $\tau$, the current curriculum stage is terminated, and the subsequent curriculum stage is initiated.
Here, the maximum patience is a fixed hyper-parameter, and the detailed process and overall curriculum learning procedure are outlined in Algorithm~\ref{alg:pacing_fuc} and~\ref{alg:overall_cl} in Appendix~\textcolor{red}{D}, respectively. 




\vspace{-0.1cm}
\section{Experimental Results}
\label{sec:exp}
\vspace{-0.2cm}
In this section, we present experimental results to validate the effectiveness of our method.
The advantages of our curriculum method, \textbf{1) Improved Performance}, \textbf{2) Faster Convergence}, and \textbf{3) Orthogonality with Existing Improvements}, are validated in this section.
To begin, we outline our experimental setups in Section~\ref{sec:exp:setup}.
Then, we provide the results of the comparative evaluation in Section~\ref{sec:exp:comparative_eval}, showing that our curriculum approach significantly improves the quality of generated samples compared to the baseline.
Finally, analyses of our method are illustrated in Section~\ref{sec:exp_analysis} to deeply understand the effectiveness of our method.

\subsection{Experimental Setup}
\label{sec:exp:setup}
Here, we provide experimental setups concisely. Detailed setups are presented in Appendix~\textcolor{red}{E}. 

\paragraph{Evaluation protocols.} 
For our comprehensive evaluation of various methods, we employed three distinct image-generation tasks:
\textbf{1) Unconditional generation} with the FFHQ dataset~\citep{karras2019style}, \textbf{2) Class-conditional generation} with CIFAR-10~\citep{krizhevsky2009learning} and ImageNet~\citep{deng2009imagenet} datasets, and \textbf{3) Text-to-Image generation} with MS-COCO dataset~\citep{lin2014microsoft}.
In 2) and 3) setups, we applied classifier-free guidance~\citep{ho2022classifier}.



\paragraph{Target models.}
We employed three exemplary diffusion architectures for experiments: DiT~\citep{peebles2022scalable}, which integrates latent diffusion models~\citep{rombach2022high} with Transformer architectures~\citep{vaswani2017attention} parameterized as $\epsilon$-prediction, EDM~\citep{karras2022elucidating}, which focuses on pixel-level diffusion utilizing UNet-based architectures~\citep{ronneberger2015u} parameterized as $F$-prediction, and SiT~\citep{ma2024sit} for score- and velocity-prediction.
For the text-to-image generation, we incorporated a CLIP text encoder~\citep{radford2021learning} as described in DTR~\citep{park2024denoising}.


\subsection{Comparative Results}
\label{sec:exp:comparative_eval}

In this section, we assess the effectiveness of our curriculum-based training approach. For a thorough comparison, we examine three distinct training variants, with further details provided in Appendix~\textcolor{red}{E}:
To achieve this, we compare three variants of training: 
\textit{1) Vanilla}: This term refers to diffusion models trained using conventional methods without any curriculum learning strategies;
\textit{2) NaiveCL}: In this variant, we incorporate a basic curriculum learning strategy, which simply repeats the same number of iterations for each stage across an $N$-stage process and does not employ SNR-based clustering; 
\textit{3) Ours}: This denotes our proposed curriculum approach, which is designed to enhance the training process of diffusion models by systematically structuring the learning stages.

\paragraph{Quantitative evaluation.}
\begin{table*}[t]
\caption{We evaluated unconditional image generation on FFHQ with DiT-B, EDM, and SiT-B, class-conditional image generation on ImageNet and CIFAR10 with DiT-L and EDM, respectively, and text-conditional image generation on MS-COCO with DiT-B. Note that our curriculum learning for diffusion models improves substantial performance without any additional parameters.} 
\vspace{-4mm}
    \begin{center}
    \begin{small}
    \scalebox{0.9}{ 
    \begin{tabular}{lcccccc}
    \toprule
    \multicolumn{7}{l}{\bf{\normalsize $\epsilon$-prediction} } \\
    \toprule
    \multirow{2}{*}{Model} & FFHQ 256$\times$256 & \multicolumn{4}{c}{ImageNet 256$\times$256} & COCO 256$\times$256  \\
    \arrayrulecolor{gray}\cmidrule(lr){2-2} \cmidrule(lr){3-6} \cmidrule(lr){7-7} 
    & FID$\downarrow$ & FID$\downarrow$ & IS$\uparrow$ & Prec$\uparrow$ & Rec$\uparrow$ & FID$\downarrow$\\
    \arrayrulecolor{black}\midrule 
    DiT (Vanilla)     &   10.49    & 11.18 & 146.95 & 0.75 & 0.47 & 7.62   \\
    \arrayrulecolor{gray}\cmidrule(lr){1-7}
    DiT + NaiveCL        & 7.95  & 11.90 & 151.66 & 0.75 & 0.47 & 7.71 \\
    \textbf{DiT + Ours}       & \textbf{7.55} & \textbf{8.18} & \textbf{186.37} & \textbf{0.79} & {0.47} & \textbf{7.51} \\
    \arrayrulecolor{black}\bottomrule
    \end{tabular}
    }
    \scalebox{0.75}{ 
    \begin{tabular}{lccccc}
    \toprule
    \multicolumn{3}{l}{\bf{\normalsize $F$-prediction} } \\
    \toprule
    \multirow{2}{*}{Model}  & FFHQ 64$\times$64 & CIFAR10 32$\times$32 \\
    \arrayrulecolor{gray}\cmidrule(lr){2-2} \cmidrule(lr){3-3}
    & FID$\downarrow$ & FID$\downarrow$ \\
    \arrayrulecolor{black}\midrule 
    EDM (Vanilla)            & 2.93 &  2.67 \\
    \arrayrulecolor{gray}\cmidrule(lr){1-6}
    EDM + NaiveCL        &  3.13 & 2.88 \\
    \textbf{EDM + Ours}       &  \textbf{2.71} & \textbf{2.44} \\
    \arrayrulecolor{black}\bottomrule
    \end{tabular}
    }
    \scalebox{0.75}{ 
    \begin{tabular}{lcccc}
    \toprule
    \multicolumn{3}{l}{\bf{\normalsize Velocity-prediction} } \\
    \toprule
    \multirow{2}{*}{Model}  & FFHQ 256$\times$256  \\
    \arrayrulecolor{gray}\cmidrule(lr){2-2}
    & FID$\downarrow$  \\
    \arrayrulecolor{black}\midrule 
    SiT (Vanilla)            &  7.44 \\
    \arrayrulecolor{gray}\cmidrule(lr){1-5}
    SiT + NaiveCL        &  7.69  \\
    \textbf{SiT + Ours}       &  \textbf{6.95}  \\
    \arrayrulecolor{black}\bottomrule
    \end{tabular}
    }
    \scalebox{0.75}{ 
    \begin{tabular}{lcccc}
    \toprule
    \multicolumn{3}{l}{\bf{\normalsize Score-prediction} } \\
    \toprule
    \multirow{2}{*}{Model}  & FFHQ 256$\times$256  \\
    \arrayrulecolor{gray}\cmidrule(lr){2-2}
    & FID$\downarrow$  \\
    \arrayrulecolor{black}\midrule 
    SiT (Vanilla)            &  9.64 \\
    \arrayrulecolor{gray}\cmidrule(lr){1-5}
    SiT + NaiveCL        &  9.77  \\
    \textbf{SiT + Ours}       &  \textbf{9.15}  \\
    \arrayrulecolor{black}\bottomrule
    \end{tabular}
    }
    \end{small}
    \end{center}
    \vspace{-7mm}
\label{tab:fid}
\end{table*}
\begin{wraptable}[10]{r}{0.39\textwidth}
    \centering
    \vspace{-0.5cm}
    \caption{Evaluating the effectiveness of curriculum learning with extended training iterations on the ImageNet 256x256 dataset using the DiT-L architecture.}
    \vspace{-0.3cm}
    \begin{tabular}{c|c|c}
    \toprule
         \diagbox[height=1cm, width=3.5cm]{Method}{FID$\downarrow$}{Iteration} & 400k & \textbf{2M} \\ \toprule
         DiT (Vanilla) & 11.18 & 7.84 \\
         \textbf{DiT + Ours} & \textbf{8.18} & \textbf{6.24} \\
    \bottomrule
    \end{tabular}
    \label{tab:2m}
    \vspace{-0.5cm}
\end{wraptable}
We quantitatively validate the effectiveness of our methods with various architectures-DiT~\citep{peebles2022scalable}, EDM~\citep{karras2022elucidating}, and SiT~\citep{ma2024sit}- and tasks including unconditional, class-conditional, and text-to-image generation.
Table~\ref{tab:fid} shows the results, confirming two empirical observations: 1) \textit{NaiveCL} fails to consistently achieve improved performance compared to \textit{Vanilla}, and 2) our approach outperforms both \textit{NaiveCL} and \textit{Vanilla}.
Regarding the first observation, \textit{NaiveCL} shows inconsistent improvements due to its lack of robust adaptation on incorporating task difficulties in various task conditions.
In contrast, our method demonstrates superior performance across all scenarios by improving the clustering and pacing of curriculums. 
Consequently, our approach consistently achieves significant performance enhancements across all metrics on four datasets: FFHQ~\citep{karras2019style}, ImageNet~\citep{deng2009imagenet}, CIFAR-10~\citep{krizhevsky2009learning}, and MS-COCO~\citep{lin2014microsoft}, illustrating its effectiveness regardless of data or model used.

Showing the results of longer training might demonstrate the robustness of our method in more extended training scenarios. We trained DiT-L/2 with 2M iterations and reported the results in Table~\ref{tab:2m}. Our model consistently outperformed the baseline, demonstrating its effectiveness even in prolonged training. Therefore, our method proves to be robust and effective for longer training durations.

\paragraph{Qualitative evaluation.}
Due to space constraints, we illustrate a detailed collection of generated examples in Appendix~\textcolor{red}{F}.
In summary, our curriculum methodology demonstrates a notable enhancement in the quality of the images produced, when compared to \textit{NaiveCL} and \textit{Vanilla}.


\begin{wrapfigure}[9]{r}{0.2605\textwidth}
    \centering
    \vspace{-12.5mm}
    \includegraphics[width=\linewidth]{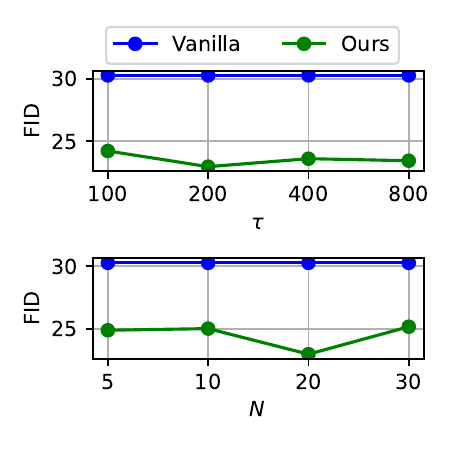}
    \vspace{-8mm}
    \caption{
    Ablation study on $N$ and $\tau$. We use DiT-B on ImageNet 256$\times$256. 
    }
    \label{fig:model_param}
\end{wrapfigure}
\subsection{Analysis}
\label{sec:exp_analysis}
To elucidate our curriculum approach's effectiveness, we present a series of analytical studies. 
All the analysis is conducted by using the DiT-B model on the ImageNet dataset.

\paragraph{Effects of $N$ and $\tau$.}
We examined the robustness of the proposed curriculum training with respect to hyper-parameters: the number of clusters $N$ and the maximum patience $\tau$. 
As shown in Fig.~\ref{fig:model_param}, our method consistently outperforms the vanilla model, and the best result is observed at $N=20, \tau=200$. 
It shows that as $\tau$ increases, it may lead to overtraining due to excessive iterations for each task, whereas as $\tau$ decreases, curriculums may not be sufficiently trained. 
Furthermore, when the entire range of timesteps is finely partitioned (i.e., with an increase in $N$), each cluster becomes excessively granular, resulting in suboptimal performance. 
Conversely, with a decrease in $N$, tasks that should be in distinct clusters are learned together, forming a coarser cluster, which also leads to suboptimal outcomes.
Overall, our method outperforms vanilla training across a range of hyperparameters, demonstrating the robustness of our approach.


\paragraph{Effects of Curriculum Design.}
\begin{wraptable}[10]{r}{0.53\textwidth}
\vspace{-4mm}
    \caption{Comparative results on various curriculum designs. 
    }
    \vspace{-2mm}
    \centering
    \resizebox{0.53\textwidth}{!}{%

    \begin{tabular}{llcccc}
    \toprule
    \multicolumn{6}{l}{\bf{Class-Conditional ImageNet} 256$\times$256.} \\
    \toprule
    \multicolumn{2}{l}{Curriculum Design}  & FID$\downarrow$ & IS$\uparrow$ & Prec$\uparrow$ & Rec$\uparrow$  \\
    \midrule
    (a) & Vanilla  & 30.27 & 60.06 & 0.55 & 0.52 \\
    \cmidrule(lr){1-6}
    (b1) &+ anti-curriculum + uniform &31.12  &62.80  &0.55	  &0.53  \\
    (b2)    &+ anti-curriculum + SNR  & 27.74 & 68.10 & 0.58 & 0.52 \\
    \cmidrule(lr){1-6}
    (c1) &+ curriculum + uniform & 25.01 & 71.99 & 0.58 & 0.53 \\
    (c2)    &+ curriculum + SNR & \textbf{22.96} & \textbf{75.98} & \textbf{0.62} & 0.52 \\
    \bottomrule
    \end{tabular}
    }
    \label{tab:design}
\end{wraptable}

In our curriculum design, we initially partitioned the entire set of timesteps into $N$ clusters using SNR-based clustering, organizing the curriculum from easy to hard clusters.
To thoroughly assess the impact of each component, we conducted the ablation study as shown in Table~\ref{tab:design}.
Firstly, we investigated the effect of curriculum learning via comparison with an anti-curriculum approach~\citep{hacohen2019power}, which progresses from hard to easy clusters, unlike conventional curriculum learning.
While both training methods in (b2) appear to enhance performance compared to vanilla training (a), anti-curriculum training cannot consistently guarantee performance improvement concerning the curriculum design as shown in (b1).
In contrast, the proposed curriculum learning method (c1, c2) consistently exhibited performance improvement even with the uniformly partitioned clusters.
Besides, with findings that utilizing SNR-clustering was more effective, clustering with the actual changes in noise levels enhanced the curriculum learning adaptability. 

\paragraph{Visualization of curriculum.}

\begin{figure*}[t]
    \begin{minipage}[b]{0.42\linewidth}
        \centering

        \includegraphics[width=\linewidth]{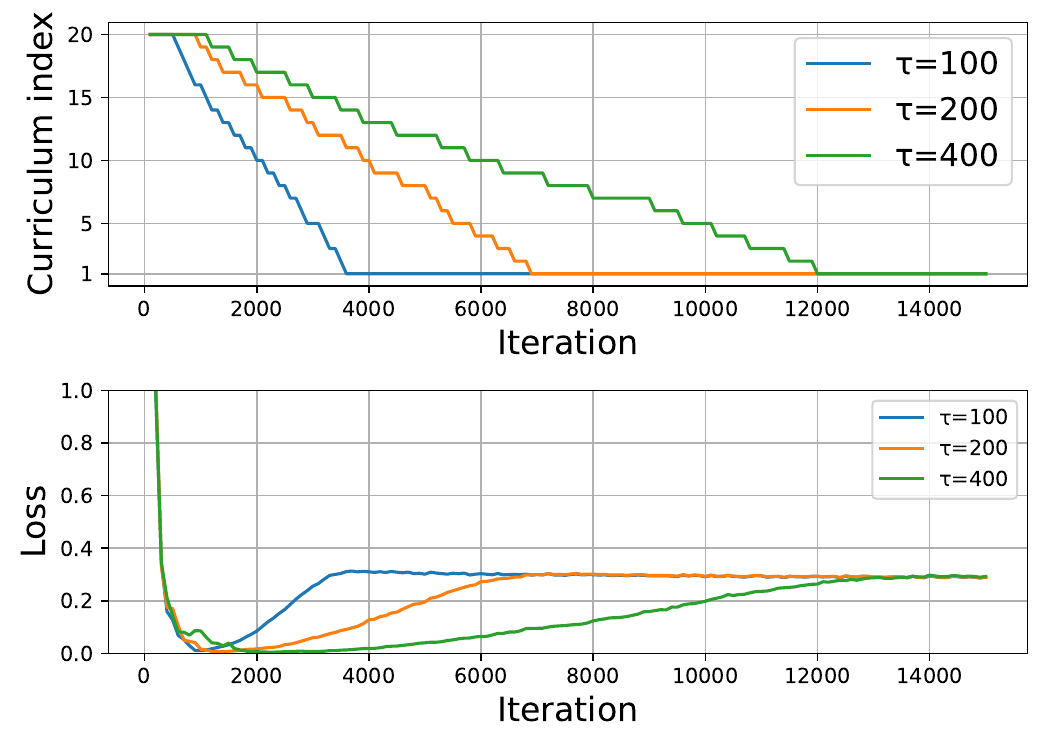}
        \vspace{-6mm}
        \caption{
        We visualized the curriculum transition and the corresponding loss across iterations ($N=20$). To make the loss graph more easily readable, the y-axis was truncated to 1.0.
        }
        \label{fig:index_loss}
    \end{minipage}%
    \hspace{1mm}
    \begin{minipage}[b]{0.56\linewidth}
        \centering
        \includegraphics[width=\linewidth]{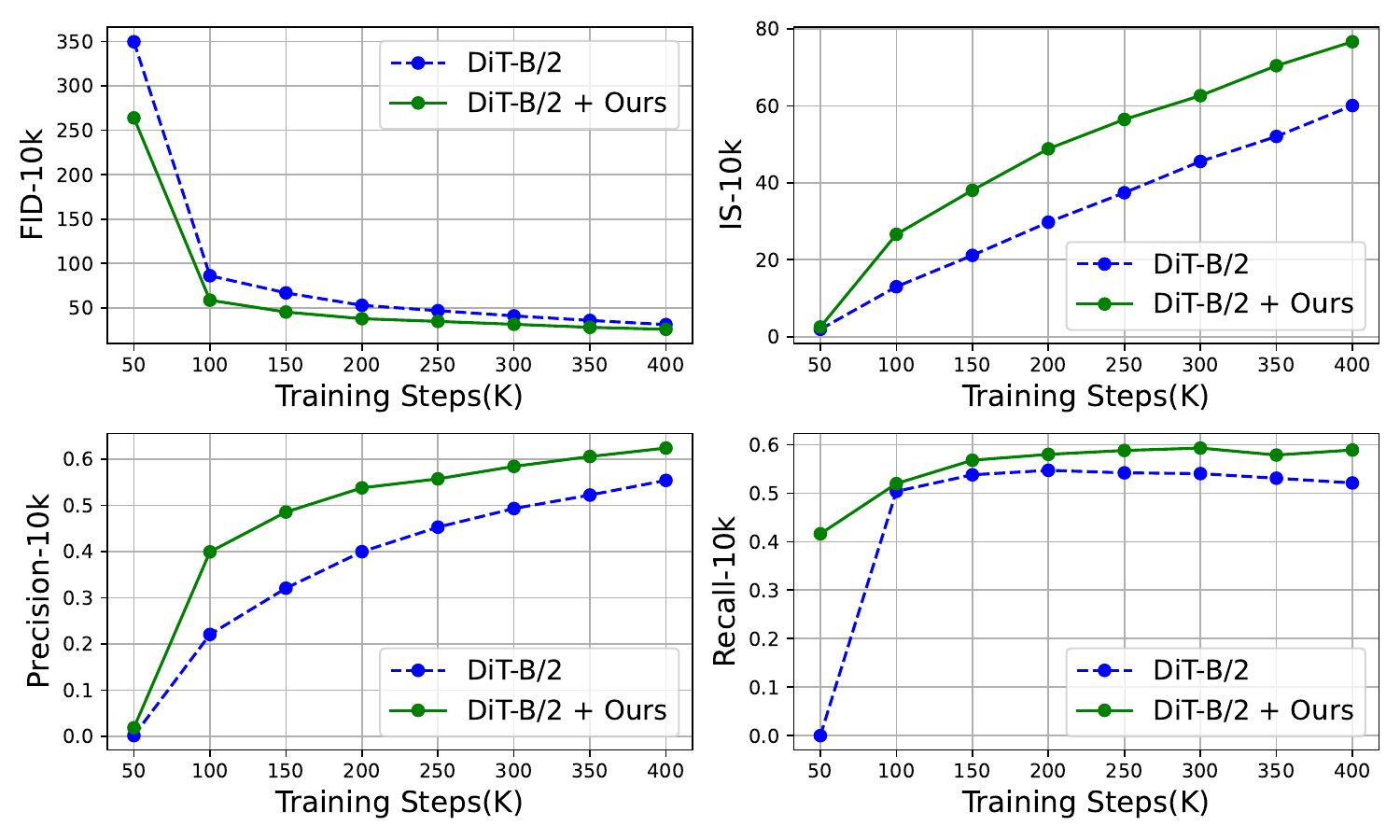}
        \vspace{-7mm}
        \caption{
        The models trained using the proposed curriculum learning approach demonstrate faster convergence compared to vanilla models, irrespective of evaluation metrics. 
        }
        \label{fig:eval_convergence}
    \end{minipage}
    \vspace{-0.15cm}
\end{figure*}

To gain deeper insights into the functioning of our curriculum pacing, we plotted loss metrics against curriculum phases, as illustrated in Fig.~\ref{fig:index_loss}. 
During the curriculum training, tasks progressively transition from the easiest to the most challenging, with varying amounts of iterations for each task based on the pacing function. 
The training loss decreased during each curriculum phase but increased after curriculum changes via the pacing function due to the inclusion of a newly added task in the updated curriculum.
Additionally, as $\tau$ increases, the curriculum phases change more gradually, highlighting the role of $\tau$ in controlling the pace of curriculum transitions.

\paragraph{Analysis on convergence speed.}
As demonstrated in previous works~\citep{bengio2009curriculum, hacohen2019power}, the adoption of curriculum learning can lead to faster convergence in model performance. 
To illustrate the efficacy of our approach in this regard, we plotted the FID, IS, precision, and recall calculated over 10,000 samples across the training iterations, as depicted in Fig.~\ref{fig:eval_convergence}. 
We observed the models trained through the proposed curriculum learning method converge faster than vanilla models, regardless of evaluation metrics. 
Notably, our approach achieves these improvements without requiring additional parameters or training iterations, thereby significantly saving time and computational resources.

\paragraph{Effectiveness on various sizes of models}
To verify the generalizability of our method across different model sizes, we evaluated the performance gains achieved using our curriculum learning approach on various scales of the DiT model: DiT-S (small), DiT-B (base), and DiT-L (large).
Table.~\ref{tab:model_size} shows that the proposed curriculum learning for diffusion model improves the performance regardless of the model size.
Moreover, it is notable that larger models exhibit a more substantial performance enhancement: DiT-S improved by 8\% in terms of FID, while DiT-B and DiT-L showed improvements of 24\% and 27\%, respectively.
These findings validate the efficacy of our curriculum approach across a diverse range of model sizes, underscoring its generalizability to various model parameters.

\paragraph{Orthogonality of Our Curriculum Approach}
Lastly, we illustrate the seamless integration of our method with sophisticated training techniques such as DTR~\citep{park2024denoising} and MinSNR~\citep{hang2023efficient}. 
Initially, we observed that each sophisticated method yields a superior performance compared to the vanilla method.
Meanwhile, as shown in Table.~\ref{tab:orthogonality}, the performance is significantly enhanced when we apply the proposed curriculum learning.
Consequently, the curriculum approach proves to be compatible with previous promising methods such as loss weighting (MinSNR) and architectural enhancements (DTR), demonstrating our orthogonality with recent diffusion techniques.

\paragraph{Additional experimental results}

Due to limited space, we present additional experimental results in Appendix~\textcolor{red}{G}.
These results also support the effectiveness of our method, emphasizing the importance of curriculum approaches in diffusion training.

\begin{table*}[t]%
    \begin{minipage}{0.35\textwidth}
    \centering
    \vspace{-0.8mm}
    \caption{Note that the curriculum learning achieves consistent improvements across the model sizes.
    }
    \vspace{-2mm}
    \resizebox{0.8\textwidth}{!}{%
    \begin{tabular}{lcccc}
    \toprule
    \multicolumn{5}{l}{\bf{Class-Conditional ImageNet} 256$\times$256.} \\
    \toprule
    Model & FID$\downarrow$  & IS$\uparrow$  & Prec$\uparrow$ & Rec$\uparrow$ \\
    \toprule
    DiT-S/2  & 43.30 & 33.63 & 0.42 & 0.54 \\
    DiT-S/2 + Ours  & \cellcolor{gray!25}\textbf{39.66} &  \cellcolor{gray!25}\textbf{36.57} & \cellcolor{gray!25}\textbf{0.44} & \cellcolor{gray!25}{0.54} \\
    \arrayrulecolor{gray}\cmidrule(lr){1-5}
    DiT-B/2 & 30.27 & 60.06 & 0.55 & 0.52 \\
    DiT-B/2 + Ours & \cellcolor{gray!25}\textbf{22.96} & \cellcolor{gray!25}\textbf{75.98} & \cellcolor{gray!25}\textbf{0.62} & \cellcolor{gray!25}{0.52} \\
    \arrayrulecolor{gray}\cmidrule(lr){1-5}
    DiT-L/2 & 11.18 & 146.95 & 0.75 & 0.47 \\
    DiT-L/2 + Ours & \cellcolor{gray!25}\textbf{8.18} & \cellcolor{gray!25}\textbf{186.37} & \cellcolor{gray!25}\textbf{0.79} & \cellcolor{gray!25}{0.47} \\
    \arrayrulecolor{gray}\cmidrule(lr){1-5}
    DiT-XL/2 & 9.40 & 166.83 & 0.77 & \textbf{0.49} \\
    DiT-XL/2 + Ours & \cellcolor{gray!25}\textbf{7.57} & \cellcolor{gray!25}\textbf{234.93} & \cellcolor{gray!25}\textbf{0.82} & \cellcolor{gray!25}{0.48} \\
    \arrayrulecolor{black}\bottomrule
    \end{tabular}%
    }
    \label{tab:model_size}
    \end{minipage}%
    \hspace{1mm}
    \begin{minipage}{0.64\textwidth}
        \centering
        \vspace{-5mm}
        \caption{Note that the curriculum learning is compatible with the previous works such as the loss weighting (MinSNR) and architecture (DTR) study which, specified the multi-task learning for diffusion model.
        } %
        \begin{small}
        \scalebox{0.65}{
        \begin{tabular}{lcccccccc}
        \toprule
        \multicolumn{9}{l}{\bf{Class-Conditional ImageNet} 256$\times$256.} \\
        \toprule
        \multirow{2}{*}{}  & \multicolumn{4}{c}{DiT-B/2} & \multicolumn{4}{c}{DiT-B/2 + Ours} \\
        \arrayrulecolor{gray}\cmidrule(lr){2-5} \cmidrule(lr){6-9}
        & FID$\downarrow$  & IS$\uparrow$ & Prec$\uparrow$ & Rec$\uparrow$ & FID$\downarrow$  & IS$\uparrow$  & Prec$\uparrow$ & Rec$\uparrow$ \\
        \arrayrulecolor{black}\toprule
        Vanilla                         & 30.27 & 60.06 & 0.55 & 0.52 & \cellcolor{gray!25}\textbf{22.96} & \cellcolor{gray!25}\textbf{75.98} & \cellcolor{gray!25}\textbf{0.62} & \cellcolor{gray!25}{0.52} \\
        MinSNR~\citep{hang2023efficient}  &21.88	&88.12 &0.63  &0.49  & \cellcolor{gray!25}\textbf{19.36} & \cellcolor{gray!25}\textbf{101.35} & \cellcolor{gray!25}\textbf{0.67} & \cellcolor{gray!25}{0.49} \\
        DTR~\citep{park2024denoising} & 15.77 & 89.89 & 0.68 & 0.52 & \cellcolor{gray!25}\textbf{15.33} & \cellcolor{gray!25}\textbf{91.39} & \cellcolor{gray!25}{0.68} & \cellcolor{gray!25}{0.52} \\
        \bottomrule
        \end{tabular}
        } %
        \end{small}
        \label{tab:orthogonality}
    \end{minipage} 
\end{table*}

\vspace{-0.1cm}
\section{Conclusion}
\vspace{-0.2cm}
In this study, we tackle the challenge of denoising task difficulty within the diffusion model framework and introduce a novel task difficulty-based curriculum learning approach. 
To the best of our knowledge, we are the first to define task difficulty by considering both the convergence rates of loss and performance metrics. 
Moreover, in terms of data distribution analysis, we observe a reduction in relative entropy between consecutive probability distributions as timesteps progress. 
We believe that these observations might help reorganize the conflicts of previous works regarding denoising task difficulties.
Building upon these insights, we propose a curriculum learning framework for diffusion models, comprising curriculum design and pacing strategies. 
Our experimental results convincingly demonstrate the efficacy of our approach across diverse diffusion model designs, datasets, and tasks. 
From these results, we emphasize that considering an order of learning denoising tasks is also a potential direction to improve training of diffusion models.
In future research, for further enhancements, more advanced curriculum learning strategies such as self-pacing can be elaborated.

\bibliographystyle{iclr2025_conference}
\bibliography{main}

\begin{thebibliography}{72}
\providecommand{\natexlab}[1]{#1}
\providecommand{\url}[1]{\texttt{#1}}
\expandafter\ifx\csname urlstyle\endcsname\relax
  \providecommand{\doi}[1]{doi: #1}\else
  \providecommand{\doi}{doi: \begingroup \urlstyle{rm}\Url}\fi

\bibitem[Allgower \& Georg(2003)Allgower and Georg]{allgower2003introduction}
Eugene~L Allgower and Kurt Georg.
\newblock \emph{Introduction to numerical continuation methods}.
\newblock SIAM, 2003.

\bibitem[Balaji et~al.(2022)Balaji, Nah, Huang, Vahdat, Song, Kreis, Aittala, Aila, Laine, Catanzaro, et~al.]{balaji2022ediffi}
Yogesh Balaji, Seungjun Nah, Xun Huang, Arash Vahdat, Jiaming Song, Karsten Kreis, Miika Aittala, Timo Aila, Samuli Laine, Bryan Catanzaro, et~al.
\newblock ediffi: Text-to-image diffusion models with an ensemble of expert denoisers.
\newblock \emph{arXiv preprint arXiv:2211.01324}, 2022.

\bibitem[Bengio et~al.(2009)Bengio, Louradour, Collobert, and Weston]{bengio2009curriculum}
Yoshua Bengio, J{\'e}r{\^o}me Louradour, Ronan Collobert, and Jason Weston.
\newblock Curriculum learning.
\newblock In \emph{Proceedings of the 26th annual international conference on machine learning}, pp.\  41--48, 2009.

\bibitem[Chang et~al.(2021)Chang, Yeh, and Demberg]{chang-etal-2021-order}
Ernie Chang, Hui-Syuan Yeh, and Vera Demberg.
\newblock Does the order of training samples matter? improving neural data-to-text generation with curriculum learning.
\newblock In Paola Merlo, Jorg Tiedemann, and Reut Tsarfaty (eds.), \emph{Proceedings of the 16th Conference of the European Chapter of the Association for Computational Linguistics: Main Volume}, pp.\  727--733. Association for Computational Linguistics, 2021.

\bibitem[Choi et~al.(2022)Choi, Lee, Shin, Kim, Kim, and Yoon]{choi2022perception}
Jooyoung Choi, Jungbeom Lee, Chaehun Shin, Sungwon Kim, Hyunwoo Kim, and Sungroh Yoon.
\newblock Perception prioritized training of diffusion models.
\newblock In \emph{Proceedings of the IEEE/CVF Conference on Computer Vision and Pattern Recognition}, pp.\  11472--11481, 2022.

\bibitem[Deja et~al.(2022)Deja, Kuzina, Trzcinski, and Tomczak]{deja2022analyzing}
Kamil Deja, Anna Kuzina, Tomasz Trzcinski, and Jakub Tomczak.
\newblock On analyzing generative and denoising capabilities of diffusion-based deep generative models.
\newblock \emph{Advances in Neural Information Processing Systems}, 35:\penalty0 26218--26229, 2022.

\bibitem[Deng et~al.(2009)Deng, Dong, Socher, Li, Li, and Fei-Fei]{deng2009imagenet}
Jia Deng, Wei Dong, Richard Socher, Li-Jia Li, Kai Li, and Li~Fei-Fei.
\newblock Imagenet: A large-scale hierarchical image database.
\newblock In \emph{2009 IEEE conference on computer vision and pattern recognition}, pp.\  248--255. Ieee, 2009.

\bibitem[Dhariwal \& Nichol(2021)Dhariwal and Nichol]{dhariwal2021diffusion}
Prafulla Dhariwal and Alexander Nichol.
\newblock Diffusion models beat gans on image synthesis.
\newblock \emph{Advances in neural information processing systems}, 34:\penalty0 8780--8794, 2021.

\bibitem[Dockhorn et~al.(2021)Dockhorn, Vahdat, and Kreis]{dockhorn2021score}
Tim Dockhorn, Arash Vahdat, and Karsten Kreis.
\newblock Score-based generative modeling with critically-damped langevin diffusion.
\newblock \emph{arXiv preprint arXiv:2112.07068}, 2021.

\bibitem[Fifty et~al.(2021)Fifty, Amid, Zhao, Yu, Anil, and Finn]{fifty2021efficiently}
Chris Fifty, Ehsan Amid, Zhe Zhao, Tianhe Yu, Rohan Anil, and Chelsea Finn.
\newblock Efficiently identifying task groupings for multi-task learning.
\newblock \emph{Advances in Neural Information Processing Systems}, 34:\penalty0 27503--27516, 2021.

\bibitem[Glasserman(2004)]{glasserman2004monte}
Paul Glasserman.
\newblock \emph{Monte Carlo methods in financial engineering}, volume~53.
\newblock Springer, 2004.

\bibitem[Go et~al.(2023{\natexlab{a}})Go, Kim, Lee, Lee, Oh, Moon, and Choi]{go2023addressing}
Hyojun Go, JinYoung Kim, Yunsung Lee, Seunghyun Lee, Shinhyeok Oh, Hyeongdon Moon, and Seungtaek Choi.
\newblock Addressing negative transfer in diffusion models.
\newblock \emph{arXiv preprint arXiv:2306.00354}, 2023{\natexlab{a}}.

\bibitem[Go et~al.(2023{\natexlab{b}})Go, Lee, Kim, Lee, Jeong, Lee, and Choi]{go2023towards}
Hyojun Go, Yunsung Lee, Jin-Young Kim, Seunghyun Lee, Myeongho Jeong, Hyun~Seung Lee, and Seungtaek Choi.
\newblock Towards practical plug-and-play diffusion models.
\newblock In \emph{Proceedings of the IEEE/CVF Conference on Computer Vision and Pattern Recognition}, pp.\  1962--1971, 2023{\natexlab{b}}.

\bibitem[Hacohen \& Weinshall(2019)Hacohen and Weinshall]{hacohen2019power}
Guy Hacohen and Daphna Weinshall.
\newblock On the power of curriculum learning in training deep networks.
\newblock In \emph{International conference on machine learning}, pp.\  2535--2544. PMLR, 2019.

\bibitem[Hang et~al.(2023)Hang, Gu, Li, Bao, Chen, Hu, Geng, and Guo]{hang2023efficient}
Tiankai Hang, Shuyang Gu, Chen Li, Jianmin Bao, Dong Chen, Han Hu, Xin Geng, and Baining Guo.
\newblock Efficient diffusion training via min-snr weighting strategy.
\newblock \emph{arXiv preprint arXiv:2303.09556}, 2023.

\bibitem[Harvey et~al.(2022)Harvey, Naderiparizi, Masrani, Weilbach, and Wood]{harvey2022flexible}
William Harvey, Saeid Naderiparizi, Vaden Masrani, Christian Weilbach, and Frank Wood.
\newblock Flexible diffusion modeling of long videos.
\newblock \emph{Advances in Neural Information Processing Systems}, 35:\penalty0 27953--27965, 2022.

\bibitem[Heusel et~al.(2017)Heusel, Ramsauer, Unterthiner, Nessler, and Hochreiter]{heusel2017gans}
Martin Heusel, Hubert Ramsauer, Thomas Unterthiner, Bernhard Nessler, and Sepp Hochreiter.
\newblock Gans trained by a two time-scale update rule converge to a local nash equilibrium.
\newblock \emph{Advances in neural information processing systems}, 30, 2017.

\bibitem[Ho \& Salimans(2022)Ho and Salimans]{ho2022classifier}
Jonathan Ho and Tim Salimans.
\newblock Classifier-free diffusion guidance.
\newblock \emph{arXiv preprint arXiv:2207.12598}, 2022.

\bibitem[Ho et~al.(2020)Ho, Jain, and Abbeel]{ho2020denoising}
Jonathan Ho, Ajay Jain, and Pieter Abbeel.
\newblock Denoising diffusion probabilistic models.
\newblock \emph{Advances in neural information processing systems}, 33:\penalty0 6840--6851, 2020.

\bibitem[Ho et~al.(2022{\natexlab{a}})Ho, Chan, Saharia, Whang, Gao, Gritsenko, Kingma, Poole, Norouzi, Fleet, et~al.]{ho2022imagen}
Jonathan Ho, William Chan, Chitwan Saharia, Jay Whang, Ruiqi Gao, Alexey Gritsenko, Diederik~P Kingma, Ben Poole, Mohammad Norouzi, David~J Fleet, et~al.
\newblock Imagen video: High definition video generation with diffusion models.
\newblock \emph{arXiv preprint arXiv:2210.02303}, 2022{\natexlab{a}}.

\bibitem[Ho et~al.(2022{\natexlab{b}})Ho, Saharia, Chan, Fleet, Norouzi, and Salimans]{ho2022cascaded}
Jonathan Ho, Chitwan Saharia, William Chan, David~J Fleet, Mohammad Norouzi, and Tim Salimans.
\newblock Cascaded diffusion models for high fidelity image generation.
\newblock \emph{Journal of Machine Learning Research}, 23\penalty0 (47):\penalty0 1--33, 2022{\natexlab{b}}.

\bibitem[Jabri et~al.(2022)Jabri, Fleet, and Chen]{jabri2022scalable}
Allan Jabri, David Fleet, and Ting Chen.
\newblock Scalable adaptive computation for iterative generation.
\newblock \emph{arXiv preprint arXiv:2212.11972}, 2022.

\bibitem[Jiang et~al.(2014)Jiang, Meng, Yu, Lan, Shan, and Hauptmann]{jiang2014self}
Lu~Jiang, Deyu Meng, Shoou-I Yu, Zhenzhong Lan, Shiguang Shan, and Alexander Hauptmann.
\newblock Self-paced learning with diversity.
\newblock \emph{Advances in neural information processing systems}, 27, 2014.

\bibitem[Karras et~al.(2019)Karras, Laine, and Aila]{karras2019style}
Tero Karras, Samuli Laine, and Timo Aila.
\newblock A style-based generator architecture for generative adversarial networks.
\newblock In \emph{Proceedings of the IEEE/CVF conference on computer vision and pattern recognition}, pp.\  4401--4410, 2019.

\bibitem[Karras et~al.(2022)Karras, Aittala, Aila, and Laine]{karras2022elucidating}
Tero Karras, Miika Aittala, Timo Aila, and Samuli Laine.
\newblock Elucidating the design space of diffusion-based generative models.
\newblock \emph{Advances in Neural Information Processing Systems}, 35:\penalty0 26565--26577, 2022.

\bibitem[Karras et~al.(2023)Karras, Aittala, Lehtinen, Hellsten, Aila, and Laine]{karras2023analyzing}
Tero Karras, Miika Aittala, Jaakko Lehtinen, Janne Hellsten, Timo Aila, and Samuli Laine.
\newblock Analyzing and improving the training dynamics of diffusion models.
\newblock \emph{arXiv preprint arXiv:2312.02696}, 2023.

\bibitem[Karras et~al.(2024)Karras, Aittala, Kynk{\"a}{\"a}nniemi, Lehtinen, Aila, and Laine]{karras2024guiding}
Tero Karras, Miika Aittala, Tuomas Kynk{\"a}{\"a}nniemi, Jaakko Lehtinen, Timo Aila, and Samuli Laine.
\newblock Guiding a diffusion model with a bad version of itself.
\newblock \emph{arXiv preprint arXiv:2406.02507}, 2024.

\bibitem[Kim et~al.(2022)Kim, Shin, Song, Kang, and Moon]{kim2022soft}
Dongjun Kim, Seungjae Shin, Kyungwoo Song, Wanmo Kang, and Il-Chul Moon.
\newblock Soft truncation: A universal training technique of score-based diffusion model for high precision score estimation.
\newblock In \emph{International Conference on Machine Learning}, pp.\  11201--11228. PMLR, 2022.

\bibitem[Kingma \& Gao(2023)Kingma and Gao]{kingma2024understanding}
Diederik Kingma and Ruiqi Gao.
\newblock Understanding diffusion objectives as the elbo with simple data augmentation.
\newblock \emph{Advances in Neural Information Processing Systems}, 36, 2023.

\bibitem[Kong et~al.(2021)Kong, Liu, Wang, and Tao]{kong2021adaptive}
Yajing Kong, Liu Liu, Jun Wang, and Dacheng Tao.
\newblock Adaptive curriculum learning.
\newblock In \emph{Proceedings of the IEEE/CVF International Conference on Computer Vision}, pp.\  5067--5076, 2021.

\bibitem[Krizhevsky et~al.(2009)Krizhevsky, Hinton, et~al.]{krizhevsky2009learning}
Alex Krizhevsky, Geoffrey Hinton, et~al.
\newblock Learning multiple layers of features from tiny images.
\newblock \emph{Master's thesis}, 2009.

\bibitem[Kumar et~al.(2010)Kumar, Packer, and Koller]{kumar2010self}
M~Kumar, Benjamin Packer, and Daphne Koller.
\newblock Self-paced learning for latent variable models.
\newblock \emph{Advances in neural information processing systems}, 23, 2010.

\bibitem[Kynk{\"a}{\"a}nniemi et~al.(2019)Kynk{\"a}{\"a}nniemi, Karras, Laine, Lehtinen, and Aila]{kynkaanniemi2019improved}
Tuomas Kynk{\"a}{\"a}nniemi, Tero Karras, Samuli Laine, Jaakko Lehtinen, and Timo Aila.
\newblock Improved precision and recall metric for assessing generative models.
\newblock \emph{Advances in Neural Information Processing Systems}, 32, 2019.

\bibitem[Lee et~al.(2023)Lee, Kim, Go, Jeong, Oh, and Choi]{lee2023multi}
Yunsung Lee, Jin-Young Kim, Hyojun Go, Myeongho Jeong, Shinhyeok Oh, and Seungtaek Choi.
\newblock Multi-architecture multi-expert diffusion models.
\newblock \emph{arXiv preprint arXiv:2306.04990}, 2023.

\bibitem[Li et~al.(2023)Li, Li, Zheng, Wu, Xiao, Wang, Zheng, Pan, Chao, and Ji]{li2023autodiffusion}
Lijiang Li, Huixia Li, Xiawu Zheng, Jie Wu, Xuefeng Xiao, Rui Wang, Min Zheng, Xin Pan, Fei Chao, and Rongrong Ji.
\newblock Autodiffusion: Training-free optimization of time steps and architectures for automated diffusion model acceleration.
\newblock In \emph{Proceedings of the IEEE/CVF International Conference on Computer Vision}, pp.\  7105--7114, 2023.

\bibitem[Lin et~al.(2014)Lin, Maire, Belongie, Hays, Perona, Ramanan, Doll{\'a}r, and Zitnick]{lin2014microsoft}
Tsung-Yi Lin, Michael Maire, Serge Belongie, James Hays, Pietro Perona, Deva Ramanan, Piotr Doll{\'a}r, and C~Lawrence Zitnick.
\newblock Microsoft coco: Common objects in context.
\newblock In \emph{Computer Vision--ECCV 2014: 13th European Conference, Zurich, Switzerland, September 6-12, 2014, Proceedings, Part V 13}, pp.\  740--755. Springer, 2014.

\bibitem[Liu et~al.(2023{\natexlab{a}})Liu, Ning, Lin, Yang, and Wang]{liu2023oms}
Enshu Liu, Xuefei Ning, Zinan Lin, Huazhong Yang, and Yu~Wang.
\newblock Oms-dpm: Optimizing the model schedule for diffusion probabilistic models.
\newblock \emph{arXiv preprint arXiv:2306.08860}, 2023{\natexlab{a}}.

\bibitem[Liu et~al.(2023{\natexlab{b}})Liu, Lin, Zeng, Long, Liu, Komura, and Wang]{liu2023syncdreamer}
Yuan Liu, Cheng Lin, Zijiao Zeng, Xiaoxiao Long, Lingjie Liu, Taku Komura, and Wenping Wang.
\newblock Syncdreamer: Generating multiview-consistent images from a single-view image.
\newblock \emph{arXiv preprint arXiv:2309.03453}, 2023{\natexlab{b}}.

\bibitem[Loshchilov \& Hutter(2017)Loshchilov and Hutter]{loshchilov2017decoupled}
Ilya Loshchilov and Frank Hutter.
\newblock Decoupled weight decay regularization.
\newblock \emph{arXiv preprint arXiv:1711.05101}, 2017.

\bibitem[Lu et~al.(2022)Lu, Zhou, Bao, Chen, Li, and Zhu]{lu2022dpm}
Cheng Lu, Yuhao Zhou, Fan Bao, Jianfei Chen, Chongxuan Li, and Jun Zhu.
\newblock Dpm-solver: A fast ode solver for diffusion probabilistic model sampling in around 10 steps.
\newblock \emph{Advances in Neural Information Processing Systems}, 35:\penalty0 5775--5787, 2022.

\bibitem[Ma et~al.(2024)Ma, Goldstein, Albergo, Boffi, Vanden-Eijnden, and Xie]{ma2024sit}
Nanye Ma, Mark Goldstein, Michael~S Albergo, Nicholas~M Boffi, Eric Vanden-Eijnden, and Saining Xie.
\newblock Sit: Exploring flow and diffusion-based generative models with scalable interpolant transformers.
\newblock \emph{arXiv preprint arXiv:2401.08740}, 2024.

\bibitem[McLeish(2011)]{mcleish2011general}
Don McLeish.
\newblock A general method for debiasing a monte carlo estimator.
\newblock \emph{Monte Carlo methods and applications}, 17\penalty0 (4):\penalty0 301--315, 2011.

\bibitem[Nichol \& Dhariwal(2021)Nichol and Dhariwal]{nichol2021improved}
Alexander~Quinn Nichol and Prafulla Dhariwal.
\newblock Improved denoising diffusion probabilistic models.
\newblock In \emph{International Conference on Machine Learning}, pp.\  8162--8171. PMLR, 2021.

\bibitem[Pan et~al.(2024)Pan, Zhuang, Huang, Nie, Yu, Xiao, Cai, and Anandkumar]{pan2023t}
Zizheng Pan, Bohan Zhuang, De-An Huang, Weili Nie, Zhiding Yu, Chaowei Xiao, Jianfei Cai, and Anima Anandkumar.
\newblock T-stitch: Accelerating sampling in pre-trained diffusion models with trajectory stitching.
\newblock \emph{arXiv preprint arXiv:2402.14167}, 2024.

\bibitem[Park et~al.(2024{\natexlab{a}})Park, Go, Kim, Woo, Ham, and Kim]{park2024switch}
Byeongjun Park, Hyojun Go, Jin-Young Kim, Sangmin Woo, Seokil Ham, and Changick Kim.
\newblock Switch diffusion transformer: Synergizing denoising tasks with sparse mixture-of-experts.
\newblock \emph{arXiv preprint arXiv:2403.09176}, 2024{\natexlab{a}}.

\bibitem[Park et~al.(2024{\natexlab{b}})Park, Woo, Go, Kim, and Kim]{park2024denoising}
Byeongjun Park, Sangmin Woo, Hyojun Go, Jin-Young Kim, and Changick Kim.
\newblock Denoising task routing for diffusion models.
\newblock In \emph{The Twelfth International Conference on Learning Representations}, 2024{\natexlab{b}}.

\bibitem[Peebles \& Xie(2022)Peebles and Xie]{peebles2022scalable}
William Peebles and Saining Xie.
\newblock Scalable diffusion models with transformers.
\newblock \emph{arXiv preprint arXiv:2212.09748}, 2022.

\bibitem[Pentina et~al.(2015)Pentina, Sharmanska, and Lampert]{pentina2015curriculum}
Anastasia Pentina, Viktoriia Sharmanska, and Christoph~H Lampert.
\newblock Curriculum learning of multiple tasks.
\newblock In \emph{Proceedings of the IEEE conference on computer vision and pattern recognition}, pp.\  5492--5500, 2015.

\bibitem[Radford et~al.(2021)Radford, Kim, Hallacy, Ramesh, Goh, Agarwal, Sastry, Askell, Mishkin, Clark, et~al.]{radford2021learning}
Alec Radford, Jong~Wook Kim, Chris Hallacy, Aditya Ramesh, Gabriel Goh, Sandhini Agarwal, Girish Sastry, Amanda Askell, Pamela Mishkin, Jack Clark, et~al.
\newblock Learning transferable visual models from natural language supervision.
\newblock In \emph{International conference on machine learning}, pp.\  8748--8763. PMLR, 2021.

\bibitem[Ramesh et~al.(2022)Ramesh, Dhariwal, Nichol, Chu, and Chen]{ramesh2022hierarchical}
Aditya Ramesh, Prafulla Dhariwal, Alex Nichol, Casey Chu, and Mark Chen.
\newblock Hierarchical text-conditional image generation with clip latents.
\newblock \emph{arXiv preprint arXiv:2204.06125}, 1\penalty0 (2):\penalty0 3, 2022.

\bibitem[Rombach et~al.(2022)Rombach, Blattmann, Lorenz, Esser, and Ommer]{rombach2022high}
Robin Rombach, Andreas Blattmann, Dominik Lorenz, Patrick Esser, and Bj{\"o}rn Ommer.
\newblock High-resolution image synthesis with latent diffusion models.
\newblock In \emph{Proceedings of the IEEE/CVF Conference on Computer Vision and Pattern Recognition}, pp.\  10684--10695, 2022.

\bibitem[Ronneberger et~al.(2015)Ronneberger, Fischer, and Brox]{ronneberger2015u}
Olaf Ronneberger, Philipp Fischer, and Thomas Brox.
\newblock U-net: Convolutional networks for biomedical image segmentation.
\newblock In \emph{Medical Image Computing and Computer-Assisted Intervention--MICCAI 2015: 18th International Conference, Munich, Germany, October 5-9, 2015, Proceedings, Part III 18}, pp.\  234--241. Springer, 2015.

\bibitem[Ruderman \& Bialek(1993)Ruderman and Bialek]{ruderman1993statistics}
Daniel Ruderman and William Bialek.
\newblock Statistics of natural images: Scaling in the woods.
\newblock \emph{Advances in neural information processing systems}, 6, 1993.

\bibitem[Ruderman(1994)]{ruderman1994statistics}
Daniel~L Ruderman.
\newblock The statistics of natural images.
\newblock \emph{Network: computation in neural systems}, 5\penalty0 (4):\penalty0 517, 1994.

\bibitem[Salimans \& Ho(2022)Salimans and Ho]{salimans2022progressive}
Tim Salimans and Jonathan Ho.
\newblock Progressive distillation for fast sampling of diffusion models.
\newblock \emph{arXiv preprint arXiv:2202.00512}, 2022.

\bibitem[Salimans et~al.(2016)Salimans, Goodfellow, Zaremba, Cheung, Radford, and Chen]{salimans2016improved}
Tim Salimans, Ian Goodfellow, Wojciech Zaremba, Vicki Cheung, Alec Radford, and Xi~Chen.
\newblock Improved techniques for training gans.
\newblock \emph{Advances in neural information processing systems}, 29, 2016.

\bibitem[Sohl-Dickstein et~al.(2015)Sohl-Dickstein, Weiss, Maheswaranathan, and Ganguli]{sohl2015deep}
Jascha Sohl-Dickstein, Eric Weiss, Niru Maheswaranathan, and Surya Ganguli.
\newblock Deep unsupervised learning using nonequilibrium thermodynamics.
\newblock In \emph{International conference on machine learning}, pp.\  2256--2265. PMLR, 2015.

\bibitem[Song et~al.(2020)Song, Meng, and Ermon]{song2020denoising}
Jiaming Song, Chenlin Meng, and Stefano Ermon.
\newblock Denoising diffusion implicit models.
\newblock \emph{arXiv preprint arXiv:2010.02502}, 2020.

\bibitem[Song \& Dhariwal(2023)Song and Dhariwal]{song2023improved}
Yang Song and Prafulla Dhariwal.
\newblock Improved techniques for training consistency models.
\newblock \emph{arXiv preprint arXiv:2310.14189}, 2023.

\bibitem[Song \& Ermon(2019)Song and Ermon]{song2019generative}
Yang Song and Stefano Ermon.
\newblock Generative modeling by estimating gradients of the data distribution.
\newblock \emph{Advances in neural information processing systems}, 32, 2019.

\bibitem[Song et~al.(2021)Song, Sohl-Dickstein, Kingma, Kumar, Ermon, and Poole]{song2021scorebased}
Yang Song, Jascha Sohl-Dickstein, Diederik~P Kingma, Abhishek Kumar, Stefano Ermon, and Ben Poole.
\newblock Score-based generative modeling through stochastic differential equations.
\newblock In \emph{International Conference on Learning Representations}, 2021.

\bibitem[Song et~al.(2023)Song, Dhariwal, Chen, and Sutskever]{song2023consistency}
Yang Song, Prafulla Dhariwal, Mark Chen, and Ilya Sutskever.
\newblock Consistency models.
\newblock \emph{arXiv preprint arXiv:2303.01469}, 2023.

\bibitem[Spitkovsky et~al.(2010)Spitkovsky, Alshawi, and Jurafsky]{spitkovsky2010baby}
Valentin~I Spitkovsky, Hiyan Alshawi, and Dan Jurafsky.
\newblock From baby steps to leapfrog: How “less is more” in unsupervised dependency parsing.
\newblock In \emph{Human Language Technologies: The 2010 Annual Conference of the North American Chapter of the Association for Computational Linguistics}, pp.\  751--759, 2010.

\bibitem[Tang et~al.(2023)Tang, Yang, Zhu, Zeng, and Bansal]{tang2023anytoany}
Zineng Tang, Ziyi Yang, Chenguang Zhu, Michael Zeng, and Mohit Bansal.
\newblock Any-to-any generation via composable diffusion.
\newblock In \emph{Thirty-seventh Conference on Neural Information Processing Systems}, 2023.

\bibitem[Vaswani et~al.(2017)Vaswani, Shazeer, Parmar, Uszkoreit, Jones, Gomez, Kaiser, and Polosukhin]{vaswani2017attention}
Ashish Vaswani, Noam Shazeer, Niki Parmar, Jakob Uszkoreit, Llion Jones, Aidan~N Gomez, {\L}ukasz Kaiser, and Illia Polosukhin.
\newblock Attention is all you need.
\newblock \emph{Advances in neural information processing systems}, 30, 2017.

\bibitem[Wang et~al.(2020)Wang, Wu, Liu, Zhou, and Yang]{wang-etal-2020-curriculum}
Chengyi Wang, Yu~Wu, Shujie Liu, Ming Zhou, and Zhenglu Yang.
\newblock Curriculum pre-training for end-to-end speech translation.
\newblock In Dan Jurafsky, Joyce Chai, Natalie Schluter, and Joel Tetreault (eds.), \emph{Proceedings of the 58th Annual Meeting of the Association for Computational Linguistics}, pp.\  3728--3738. Association for Computational Linguistics, 2020.

\bibitem[Woo et~al.(2023)Woo, Park, Go, Kim, and Kim]{woo2023harmonyview}
Sangmin Woo, Byeongjun Park, Hyojun Go, Jin-Young Kim, and Changick Kim.
\newblock Harmonyview: Harmonizing consistency and diversity in one-image-to-3d.
\newblock \emph{arXiv preprint arXiv:2312.15980}, 2023.

\bibitem[Wu et~al.(2020)Wu, Dyer, and Neyshabur]{wu2020curricula}
Xiaoxia Wu, Ethan Dyer, and Behnam Neyshabur.
\newblock When do curricula work?
\newblock \emph{arXiv preprint arXiv:2012.03107}, 2020.

\bibitem[Xu et~al.(2023)Xu, Tong, and Jaakkola]{xu2023stable}
Yilun Xu, Shangyuan Tong, and Tommi Jaakkola.
\newblock Stable target field for reduced variance score estimation in diffusion models.
\newblock \emph{arXiv preprint arXiv:2302.00670}, 2023.

\bibitem[Yang et~al.(2023{\natexlab{a}})Yang, Yu, Wang, Wang, Weng, Zou, and Yu]{yang2023diffsound}
Dongchao Yang, Jianwei Yu, Helin Wang, Wen Wang, Chao Weng, Yuexian Zou, and Dong Yu.
\newblock Diffsound: Discrete diffusion model for text-to-sound generation, 2023{\natexlab{a}}.

\bibitem[Yang et~al.(2023{\natexlab{b}})Yang, Zhou, Feng, and Wang]{yang2023diffusion}
Xingyi Yang, Daquan Zhou, Jiashi Feng, and Xinchao Wang.
\newblock Diffusion probabilistic model made slim.
\newblock In \emph{Proceedings of the IEEE/CVF Conference on Computer Vision and Pattern Recognition}, pp.\  22552--22562, 2023{\natexlab{b}}.

\bibitem[Yue et~al.(2024)Yue, Wang, Sun, Ji, Chang, and Zhang]{yue2024exploring}
Zhongqi Yue, Jiankun Wang, Qianru Sun, Lei Ji, Eric I-Chao Chang, and Hanwang Zhang.
\newblock Exploring diffusion time-steps for unsupervised representation learning.
\newblock In \emph{The Twelfth International Conference on Learning Representations}, 2024.

\end{thebibliography}

\newpage
\appendix
\newpage
\addtocontents{toc}{\protect\setcounter{tocdepth}{2}}   
\hypersetup{linkcolor=black}
\newpage
\section*{\centering\LARGE Appendix}
\tableofcontents

\hypersetup{linkcolor=red}
\newpage

\appendix

\section{Extended Related Work}

\subsection{Analyzing Diffusion Model Behaviors in Each Timestep}
In this section, we review works related to analyzing diffusion model behaviors in each timestep but not covered in detail in Section~\textcolor{red}{2.1}.
Deja~\etal~\citep{deja2022analyzing} explore denoising during the backward diffusion process and observe that transition from denoising to generation exists in the backward process.
Go~\etal~\citep{go2023addressing} investigate the affinity between denoising tasks, showing that temporal proximal denoising tasks exhibit higher task affinity. 
Then, they also observe that simultaneously learning all denoising tasks by one model suffers from negative transfer.
They can achieve better performance than standard diffusion training by alleviating negative transfer.
Lee~\etal~\citep{lee2023multi} analyze frequency characteristics according to timesteps and observe that high-frequency components are lost as timesteps increase.
From this observation, they propose a multi-architecture multi-experts diffusion model, which utilizes multiple denoiser models specialized in each timestep interval but utilizes a transformer-like model as the timestep increases.
From observations that smaller and larger models produce similar latent noise, Pan~\etal~\citep{pan2023t} propose T-Stitch, which leverages a pre-trained smaller model at the beginning of the backward process to accelerate the sampling speed.
Xu~\etal~\citep{xu2023stable} investigate the average trace-of-covariance of training targets according to timesteps, showing that it peaks in the intermediate timesteps, causing unstable training targets.
For more stable training targets, they utilize weighted conditional scores with a reference batch.

\subsection{Easy-to-hard training Strategy}
Progressive distillation~\citep{salimans2022progressive} focuses on reducing the number of sampling steps by training the model to progressively skip more steps, while cascaded diffusion~\citep{ho2022cascaded} aims to improve sample quality by progressively increasing the image resolution during training. 
Both methods concentrate on altering the model's behavior or structure to tackle specific challenges, such as efficiency or resolution enhancement. In contrast, our work identifies trends in task difficulty across timestep-wise denoising tasks and leverages these findings to propose an easy-to-hard training scheme.
This training strategy directly addresses the order and structure of the learning process, optimizing task sequencing to enhance performance. 
This distinction emphasizes that our approach is fundamentally different from these methods, as it addresses a unique aspect of diffusion model training.

\section{Detailed Experimental Setups for Observation}

In Section \textcolor{red}{4}, we examined the difficulty of denoising tasks in terms of convergence with various models $\{\mathrm{M}\}_{i=1}^{20}$, which are trained within specific timesteps $[\frac{i-1}{20}T, \frac{i}{20}T]$ for DiT and SiT, and $[\mathrm{\Phi}^{-1}(\frac{i-1}{N}), \mathrm{\Phi}^{-1}(\frac{i}{N})]$ for EDM where $\mathrm{\Phi}^{-1}$ is the inverse cumulative distribution function of the Gaussian distribution. 
For the DiT architecture, we employed the DiT-B/2, whereas for EDM, we used the DDPM++ architecture. 
Both DiT and EDM models were trained on the FFHQ dataset, with a batch size of 256, for approximately 20,000 iterations and 4,000 kimg iterations (equivalent to processing 1 million images), respectively. 
This training was conducted until both loss and performance converged. 
As illustrated in Fig. \textcolor{red}{A}, we additionally plotted the iterations of each timestep interval when their loss values start to oscillate. 
We measured this by counting the number of times the loss value increased after the step reached 100. 
As shown in the results, losses of all timestep intervals are stabilized within 20K iterations, while the lower timesteps reach this point more slowly. 
This also suggests that the convergence speed of lower timesteps tends to exhibit a slower regime.
To examine specifically at the observation of convergence, we also analyzed the convergence speed on the ImageNet dataset. 
As shown in Fig. \textcolor{red}{B}, we obtained similar results as on the FFHQ dataset.
Configuration of training optimizers and learning rates are the same as setups in Section \textcolor{red}{6}. 
\begin{figure*}[h!]
    \centering
    \vspace{5mm}
    \includegraphics[width=0.8\linewidth]{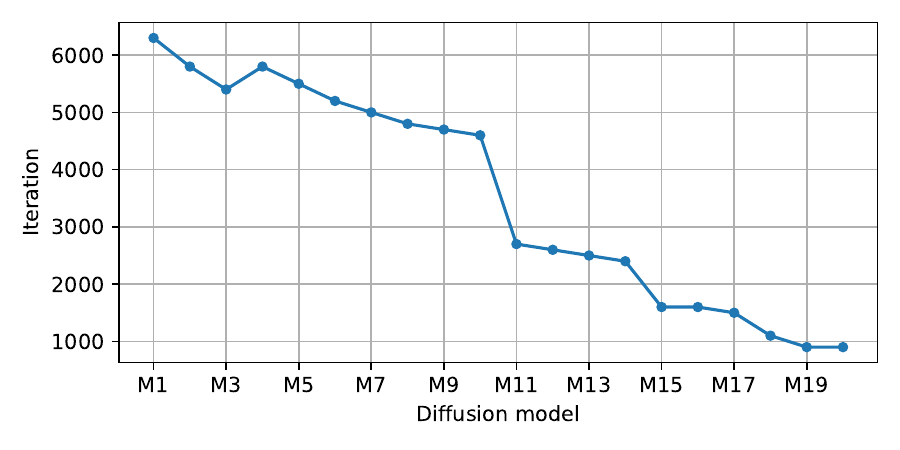}
    \vspace{-6mm}
     \caption*{\textbf{Figure A:} Converged points are plotted during training for each diffusion model $\mathrm{M}_i$ in SiT.}
     \vspace{3.5mm}
\vspace{-0.15cm}
\end{figure*}
\begin{figure*}[h!]
    \centering
    \includegraphics[width=0.8\linewidth]{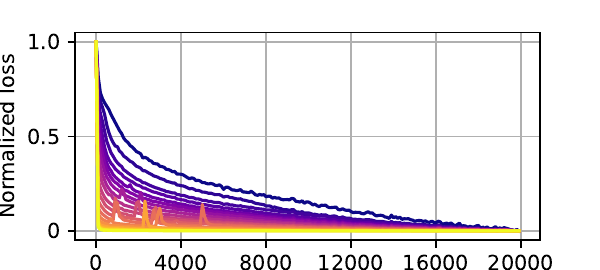}
     \caption*{\textbf{Figure B:} Loss convergence plotted during training for each diffusion model $\mathrm{M}_i$ in DiT on ImageNet dataset.}
\end{figure*}

To evaluate the performance of diffusion models through the FID score of the generated images, performing a recursive denoising task from $T$ to zero is necessary, complicating the assessment using only $\mathrm{M}_i$. 
Following~\citep{go2023addressing}, we generated samples where $\mathrm{M}_i$ was specifically utilized for denoising within its trained intervals. At the same time, the diffusion model is responsible for the denoising tasks across the entire range of timesteps. 
For this evaluation, we sampled 10K images using a DDPM sampler over 250 steps for DiT and SiT, and an Euler solver over 40 steps for the other models.

\section{Approximation of KL Divergence of $p_{t-1}$ and $p_t$.}

Here, we supplement the approximation of KL Divergence of $p_{t-1}$ and $p_t$ omitted in Section \textcolor{red}{4.2}.
To explore the difficulties of denoising tasks from the distributional viewpoint, we analyze the KL divergence of $p_{t-1}$ and $p_t$, $D_{KL}(p_{t-1}||p_t)$.
However, due to the unknown explicit density form of $p_0$, it is approximated through unbiased estimators as follows:
\begin{equation}
    \hat{D}_{KL}\left(p_{t-1} || p_t \right) = \frac{1}{M} \sum_{\substack{i\in\{1, 2, \cdots, M\} \\ \bm{x}_i \sim p_{t-1}}} \log \left( \frac{p_{t-1}(\bm{x}_i)}{p_t(\bm{x}_i)} \right),
\end{equation}
\begin{equation}
    \hat{p}_t(\bm{x}_t) = \frac{1}{L} \sum_{\substack{j \in \{1, 2, \cdots, L\} \\ \bm{y}_j \sim p_0}} p_{0t}(\bm{x}_t | \bm{x}_0=\bm{y}_j),
\end{equation}
where $\hat{D}_{KL}$ and $\hat{p}_t$ are unbiased estimators of $D_{KL}$ and $p_t$, respectively, and we choose Monte-Carlo estimators for them~\citep{glasserman2004monte, mcleish2011general}.

We sampled 5,000 images to approximate the KL divergence, which is enough for Monte-Carlo sampling and might be no changes for larger samples. Despite the large amount of samples, the exploding appearance observed in Fig. 2 when $t$ is close to zero is due to the characteristics of the data distribution. The image data distribution has narrow support (roughly speaking, it is non-zero only within a narrow range)~\citep{ruderman1993statistics,karras2024guiding}. As $t$ increases, information about the original data distribution gradually diminishes with the modes in the distribution of $x_t$ vanishing towards zero.

Given this, when $t$ is close to zero (i.e. when the distribution of $x_t$ is still analogous to the original data distribution), the narrow support and the tendency to move towards zero give rise to a region where $p_{t-1}$ does not overlap with $p_{t}$. Consequently, when calculating the KL divergence $D_{KL} (p_{t-1} || p_t) = E_{x \sim p_{t-1}} [\log (\frac{p_{t-1} ({x})}{p_t ({x})})]$, $x_{t-1}$ potentially falls outside the support of $p_{t}$, which leads to $p_{t}(x_{t-1}) = 0$ and numerical unstability. On the other hand, as $t$ increases, the accumulated noise broadens the support of $x$'s distribution, reducing the occurrence of zero values and stabilizing the numerical estimation.

\section{Algorithm}


Due to the limited space of the main manuscript, we hereby present the step-by-step process of our method to supplement the details of our approach.
The pacing function, which determines the moments to transit between curriculum stages is described in Algorithm~\ref{alg:pacing_fuc}.
By incorporating this pacing function, the detailed procedure of our proposed curriculum learning method for training diffusion is illustrated in Algorithm~\ref{alg:overall_cl}.


\begin{figure*}[ht]
    \centering
    \scalebox{0.8}{
    \begin{minipage}{.54\linewidth}
        \centering
        \begin{algorithm}[H]
        \scriptsize
        \caption{Pacing Function}
        \label{alg:pacing_fuc} 
        \begin{algorithmic}
            \State \textbf{Input:} Current loss $L_\text{cur}$, Best loss $L_\text{best}$, Current patience $\tau_\text{cur}$, Maximum patience $\tau_\text{max}$, Current curriculum index $I_\text{cur}$ \\
            \State \textbf{Output:} Updated patience, Updated curriculum index \\

            \State \mycomment{\# Reset patience}
            \vspace{0.1mm}
            \If{$L_\text{cur} < L_\text{best}$}
                \State \Return 0, $I_\text{cur}$ \hfill  \\
            \Else
                \State \mycomment{\# Proceed to next curriculum}
                \vspace{0.1mm}
                \If{$\tau_\text{cur}+ 1 > \tau_\text{max}$}
                    \State \Return 0, $I_\text{cur} - 1$ \\
                \mycomment{\quad \ \ \# Increase patience}
                \vspace{0.1mm}
                \Else
                    \State \Return $p_\text{cur}+1, I_\text{cur}$ \hfill  
                \EndIf
            \EndIf \\  
            \vspace{9.7mm}
        \end{algorithmic}
        \end{algorithm}
    \end{minipage}
    \hfill
    \scalebox{1.0}{
    \begin{minipage}{.54\linewidth}
       \centering
        \begin{algorithm}[H]
        \scriptsize
        \caption{Curriculum Learning}
        \label{alg:overall_cl} 
        \begin{algorithmic}
            \State \textbf{Input:} Curriculum $\{C_i\}_{i=1}^{N}$, Pacing function $g$,  Maximum patience $\tau_\text{max}$, Loss function $f$, Curriculum index $I_\text{cur} = N$, Best loss $L_\text{best} = \infty$, Model $M_\theta$ 
            \While{$I_\text{cur} > 0$} 
                \State \mycomment{\# Mini-batch sampling}
                \vspace{0.1mm}
                \State $X \sim C_{I_{\text{cur}}}$  
                \vspace{0.1mm}
                \State \mycomment{\# Calculate Loss}
                \vspace{0.1mm}
                \State $L_{\text{cur}} = f(M_\theta(X))$ 
                \State \mycomment{\# Update model}
                \vspace{0.1mm}
                \State $\theta$ = $\theta$ - $\nabla_\theta{L_{\text{cur}}}$ 
                \State \mycomment{\# Pacing function}
                \vspace{0.1mm}
                \State $\tau_{\text{cur}}, I_\text{next} = g(L_{\text{cur}}, L_{\text{best}}, \tau_{\text{cur}}, \tau_{\text{max}},I_{\text{cur}})$  \\
                \State \mycomment{\# Update curriculum}
                \vspace{0.1mm}
                \If{$I_\text{cur} \neq I_\text{next}$}
                    \State $I_{\text{cur}} = I_{\text{next}}$ 
                    \State $L_{\text{best}} = \infty$ \\
                \mycomment{\quad \ \ \# Update best loss}
                \vspace{0.1mm}
                \ElsIf{$L_{\text{cur}} < L_{\text{best}}$}
                    \State $L_{\text{best}} = L_{\text{cur}}$ 
                \EndIf
            \EndWhile
        \end{algorithmic}
        \end{algorithm}
    \end{minipage}
    }
    }%
\end{figure*}

\section{Details on Experimental Setups}


\paragraph{Evaluation metrics.} 
To evaluate the performance of models, we utilized three metrics: FID~\citep{heusel2017gans}, IS~\citep{salimans2016improved}, and Precision/Recall~\citep{kynkaanniemi2019improved}.
Specifically, we applied FID and IS to measure sample quality, while Precision is used to assess quality further and Recall was utilized to evaluate the diversity of the generated samples in ImageNet setup.
In other datasets, we employed FID to evaluate sample quality.
Unless otherwise mentioned, we sampled 50K samples for evaluation.
In tasks involving conditional generation, including class-conditional image generation (e.g. CIFAR-10, ImageNet) and text-to-image conversion (e.g. MS-COCO), we adapted the classifier-free guidance~\citep{ho2022classifier} with a guidance scale of 1.5.

\paragraph{Training details.}
For training diffusion models, we utilized the AdamW optimizer~\citep{loshchilov2017decoupled} with a constant learning rate of 0.0001, and weight decay was not applied. 
The exponential moving average (EMA) to the model's weights was used to stabilize the training and the decay ratio was set to 0.9999.
The batch size was set to 256, and we augmented the training data by a horizontal flip. 
While the diffusion timestep $T$ was configured as 1,000 for all experiments, we trained for 100K iterations for the FFHQ dataset~\citep{karras2019style}, and 400K iterations for the ImageNet dataset~\citep{deng2009imagenet} and MS-COCO dataset~\citep{lin2014microsoft}.
The number of clusters $N$ was 20 unless otherwise specified. The maximum patience $\tau$ was varied across model sizes: it was set at 200 for DiT-S/2, DiT-B/2, and EDM, and 400 for DiT-L/2. EDM was trained using fp16, while the other models were trained using fp32.
We used 8 A100 GPUs for all experiments.

\begin{figure*}[h]
    \centering
        \begin{tabular}{@{\hspace{1mm}}c@{\hspace{1mm}}c@{\hspace{1mm}}c@{\hspace{1mm}}c@{\hspace{1mm}}c@{\hspace{1mm}}c}

        \raisebox{0.62\height}{\rotatebox{90}{\scriptsize Vanilla}} 
        \adjincludegraphics[clip,width=0.15\textwidth,trim={0 0 0 0}]{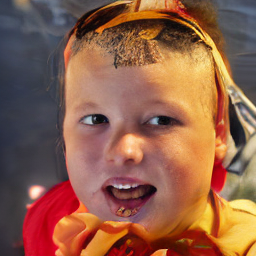} &
        \adjincludegraphics[clip,width=0.15\textwidth,trim={0 0 0 0}]{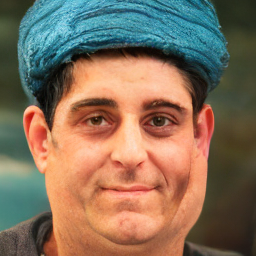} &
        \adjincludegraphics[clip,width=0.15\textwidth,trim={0 0 0 0}]{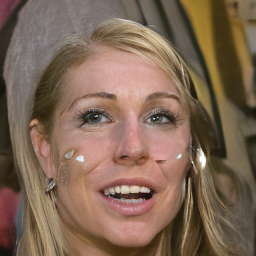} &
        \adjincludegraphics[clip,width=0.15\textwidth,trim={0 0 0 0}]{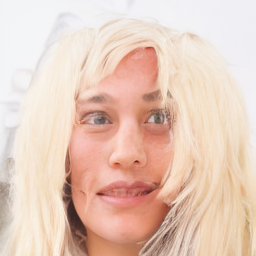} &
        \adjincludegraphics[clip,width=0.15\textwidth,trim={0 0 0 0}]{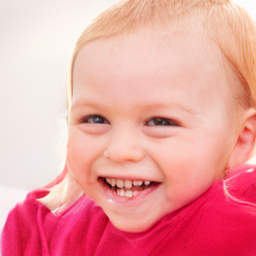}  &
        \adjincludegraphics[clip,width=0.15\textwidth,trim={0 0 0 0}]{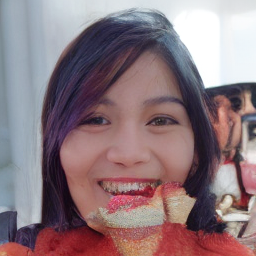}  \\ 
        
        \raisebox{0.8\height}{\rotatebox{90}{\scriptsize Naive}} 
        \adjincludegraphics[clip,width=0.15\textwidth,trim={0 0 0 0}]{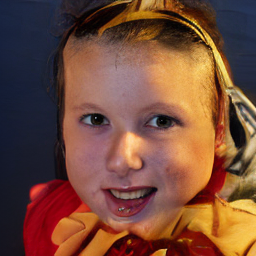} &
        \adjincludegraphics[clip,width=0.15\textwidth,trim={0 0 0 0}]{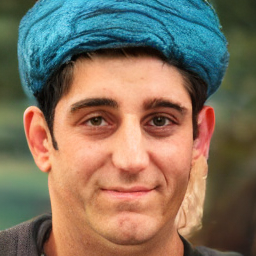} &
        \adjincludegraphics[clip,width=0.15\textwidth,trim={0 0 0 0}]{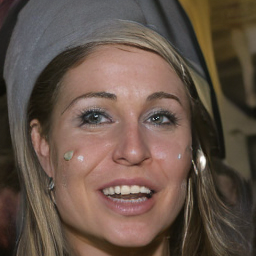} &
        \adjincludegraphics[clip,width=0.15\textwidth,trim={0 0 0 0}]{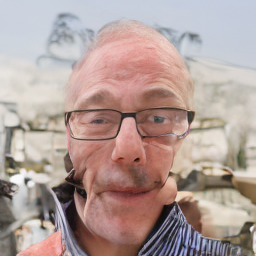} &
        \adjincludegraphics[clip,width=0.15\textwidth,trim={0 0 0 0}]{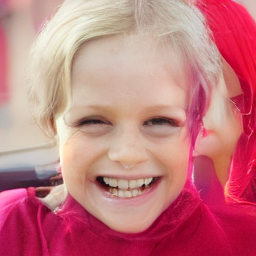}  &
        \adjincludegraphics[clip,width=0.15\textwidth,trim={0 0 0 0}]{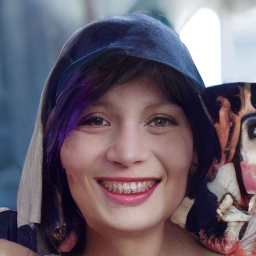}  \\
    
        \raisebox{1.1\height}{\rotatebox{90}{\scriptsize Ours}} 
        \adjincludegraphics[clip,width=0.15\textwidth,trim={0 0 0 0}]{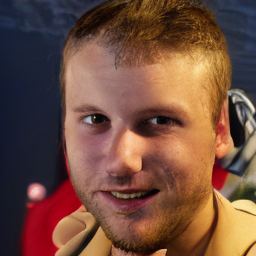} &
        \adjincludegraphics[clip,width=0.15\textwidth,trim={0 0 0 0}]{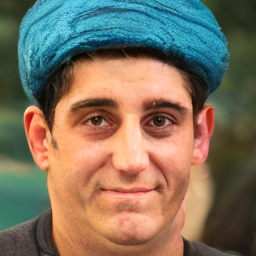} &
        \adjincludegraphics[clip,width=0.15\textwidth,trim={0 0 0 0}]{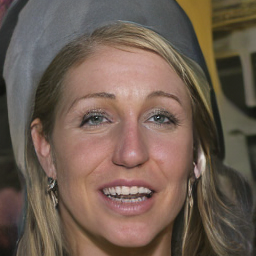} &
        \adjincludegraphics[clip,width=0.15\textwidth,trim={0 0 0 0}]{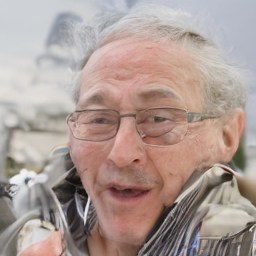} &
        \adjincludegraphics[clip,width=0.15\textwidth,trim={0 0 0 0}]{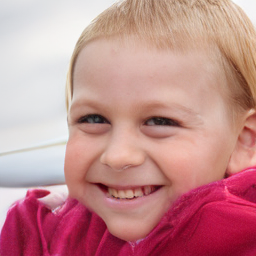} &
        \adjincludegraphics[clip,width=0.15\textwidth,trim={0 0 0 0}]{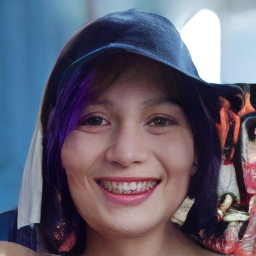}  \\
    \end{tabular}
    \caption*{\textbf{Figure C: }Qualitative comparison between vanilla, naive curriculum, and ours on the FFHQ dataset. 
    }
    \vspace{-4mm}
    \label{fig:ffhq_qualitative}

\end{figure*}

\begin{figure*}[h]
    \centering
    \begin{tabular}{@{\hspace{1mm}}c@{\hspace{1mm}}c@{\hspace{1mm}}c@{\hspace{1mm}}c@{\hspace{1mm}}c@{\hspace{1mm}}c@{\hspace{1mm}}c}
        \raisebox{0.32\height}{\rotatebox{90}{\scriptsize Vanilla}} 
        \adjincludegraphics[clip,width=0.125\textwidth,trim={0 0 0 0}]{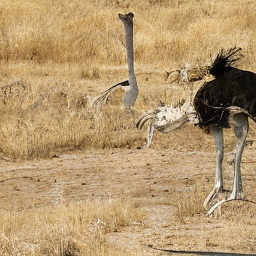} &
        \adjincludegraphics[clip,width=0.125\textwidth,trim={0 0 0 0}]{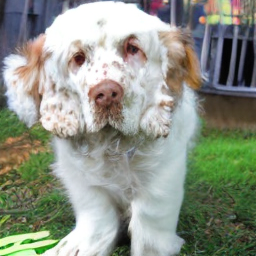} &
        \adjincludegraphics[clip,width=0.125\textwidth,trim={0 0 0 0}]{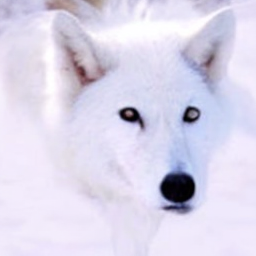} &
        \adjincludegraphics[clip,width=0.125\textwidth,trim={0 0 0 0}]{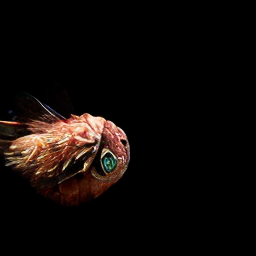} &
        \adjincludegraphics[clip,width=0.125\textwidth,trim={0 0 0 0}]{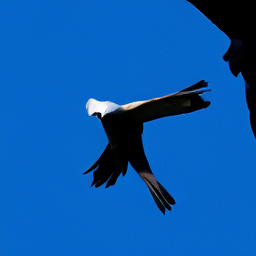} &
        \adjincludegraphics[clip,width=0.125\textwidth,trim={0 0 0 0}]{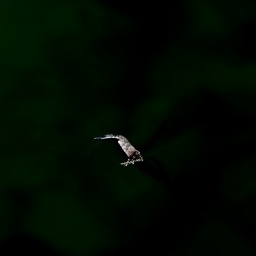} &
        \adjincludegraphics[clip,width=0.125\textwidth,trim={0 0 0 0}]{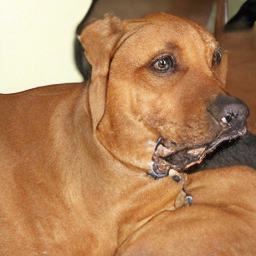} \\
        
        \raisebox{0.67\height}{\rotatebox{90}{\scriptsize Naive}} 
        \adjincludegraphics[clip,width=0.125\textwidth,trim={0 0 0 0}]{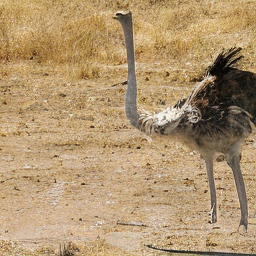} &
        \adjincludegraphics[clip,width=0.125\textwidth,trim={0 0 0 0}]{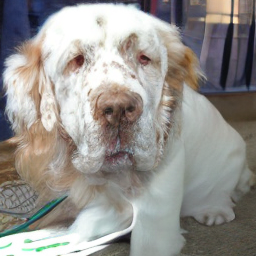} &
        \adjincludegraphics[clip,width=0.125\textwidth,trim={0 0 0 0}]{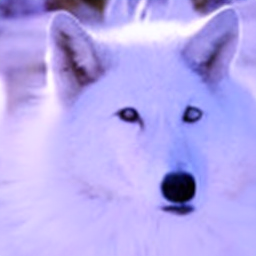} &
        \adjincludegraphics[clip,width=0.125\textwidth,trim={0 0 0 0}]{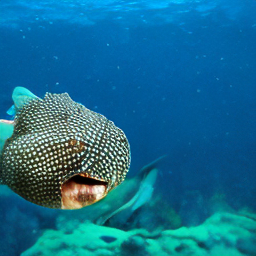} &
        \adjincludegraphics[clip,width=0.125\textwidth,trim={0 0 0 0}]{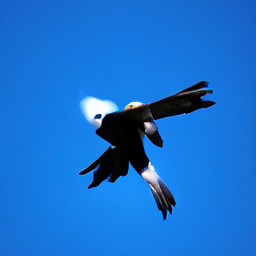} &
        \adjincludegraphics[clip,width=0.125\textwidth,trim={0 0 0 0}]{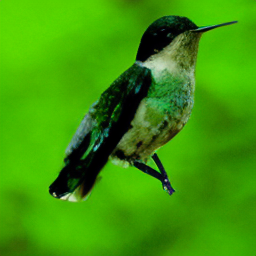} &
        \adjincludegraphics[clip,width=0.125\textwidth,trim={0 0 0 0}]{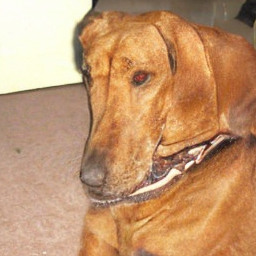} \\
        
        \raisebox{0.9\height}{\rotatebox{90}{\scriptsize Ours}} 
        \adjincludegraphics[clip,width=0.125\textwidth,trim={0 0 0 0}]{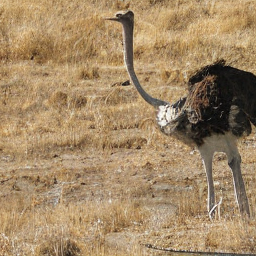} &
        \adjincludegraphics[clip,width=0.125\textwidth,trim={0 0 0 0}]{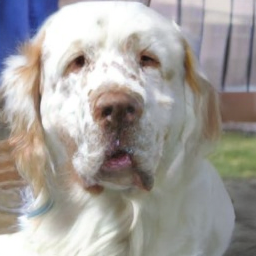} &
        \adjincludegraphics[clip,width=0.125\textwidth,trim={0 0 0 0}]{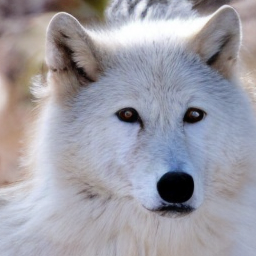} &
        \adjincludegraphics[clip,width=0.125\textwidth,trim={0 0 0 0}]{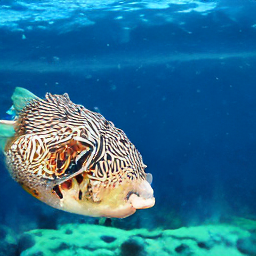} &
        \adjincludegraphics[clip,width=0.125\textwidth,trim={0 0 0 0}]{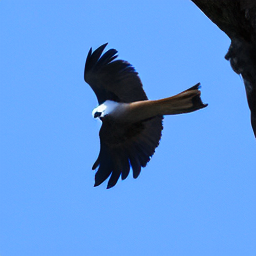} &
        \adjincludegraphics[clip,width=0.125\textwidth,trim={0 0 0 0}]{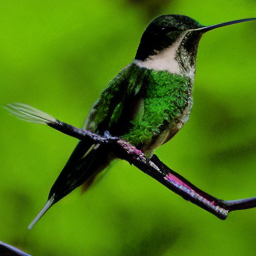} &
        \adjincludegraphics[clip,width=0.125\textwidth,trim={0 0 0 0}]{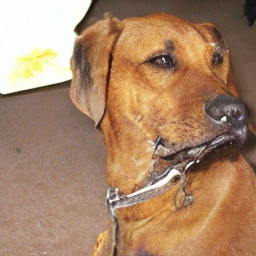} \\

        \quad{\tiny Ostrich} & {\tiny Clumber} & {\tiny Arctic wolf} & {\tiny Puffer} & {\tiny American eagle} & {\tiny Hummingbird} & {\tiny Ridgeback} \\
    \end{tabular}
    \caption*{\textbf{Figure D: }
    Qualitative comparison between vanilla, naive curriculum, and ours on ImageNet dataset.}
    \vspace{-4mm}
    \label{fig:img_qualitative}
\end{figure*}
\begin{figure*}[ht]
    \centering
        \begin{tabular}{@{\hspace{1mm}}c@{\hspace{1mm}}c@{\hspace{1mm}}c@{\hspace{1mm}}c@{\hspace{1mm}}c@{\hspace{1mm}}c@{\hspace{1mm}}c}

        \raisebox{0.57\height}{\rotatebox{90}{\scriptsize Vanilla}} 
        \adjincludegraphics[clip,width=0.13\textwidth,trim={0 0 0 0}]{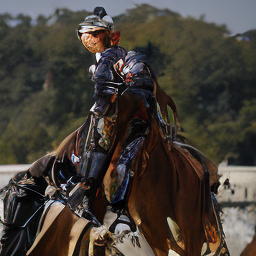} &
        \adjincludegraphics[clip,width=0.13\textwidth,trim={0 0 0 0}]{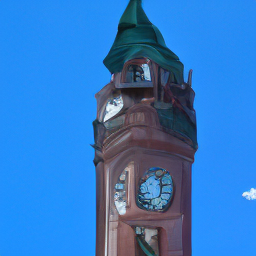} &
        \adjincludegraphics[clip,width=0.13\textwidth,trim={0 0 0 0}]{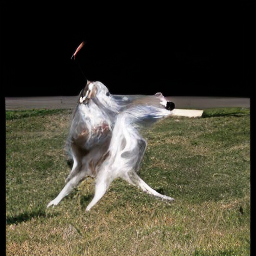} &
        \adjincludegraphics[clip,width=0.13\textwidth,trim={0 0 0 0}]{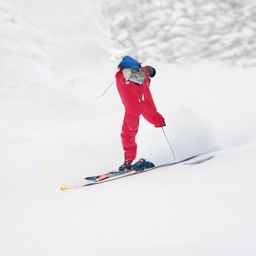} &
        \adjincludegraphics[clip,width=0.13\textwidth,trim={0 0 0 0}]{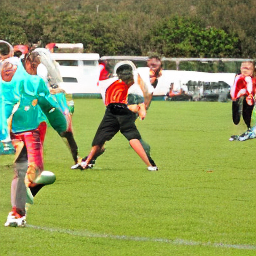}  &
        \adjincludegraphics[clip,width=0.13\textwidth,trim={0 0 0 0}]{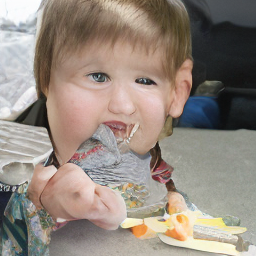}  &
        \adjincludegraphics[clip,width=0.13\textwidth,trim={0 0 0 0}]{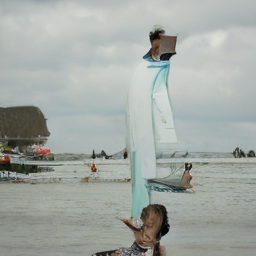}  \\ 
        
        \raisebox{0.75\height}{\rotatebox{90}{\scriptsize Naive}} 
        \adjincludegraphics[clip,width=0.13\textwidth,trim={0 0 0 0}]{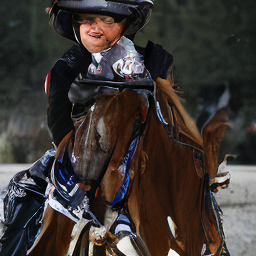} &
        \adjincludegraphics[clip,width=0.13\textwidth,trim={0 0 0 0}]{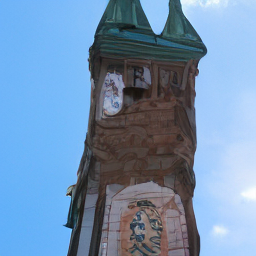} &
        \adjincludegraphics[clip,width=0.13\textwidth,trim={0 0 0 0}]{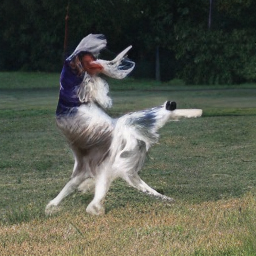} &
        \adjincludegraphics[clip,width=0.13\textwidth,trim={0 0 0 0}]{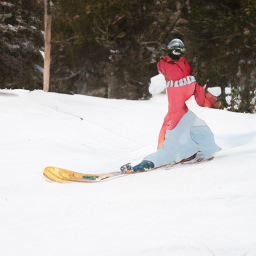} &
        \adjincludegraphics[clip,width=0.13\textwidth,trim={0 0 0 0}]{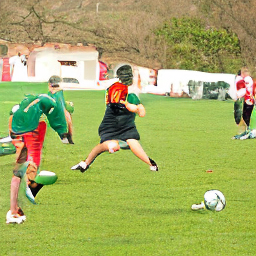}  &
        \adjincludegraphics[clip,width=0.13\textwidth,trim={0 0 0 0}]{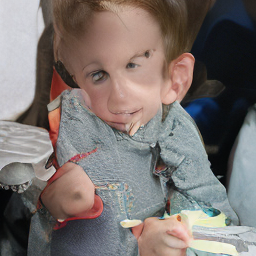}  &
        \adjincludegraphics[clip,width=0.13\textwidth,trim={0 0 0 0}]{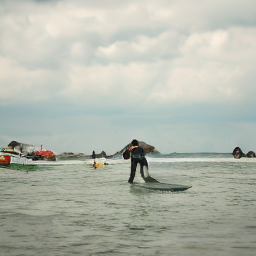}  \\
    
        \raisebox{1.1\height}{\rotatebox{90}{\scriptsize Ours}} 
        \adjincludegraphics[clip,width=0.13\textwidth,trim={0 0 0 0}]{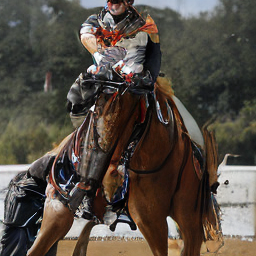} &
        \adjincludegraphics[clip,width=0.13\textwidth,trim={0 0 0 0}]{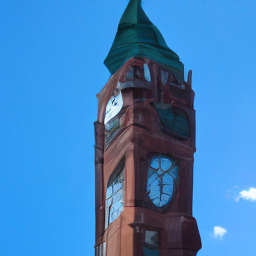} &
        \adjincludegraphics[clip,width=0.13\textwidth,trim={0 0 0 0}]{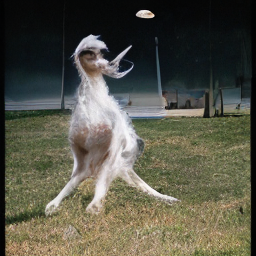} &
        \adjincludegraphics[clip,width=0.13\textwidth,trim={0 0 0 0}]{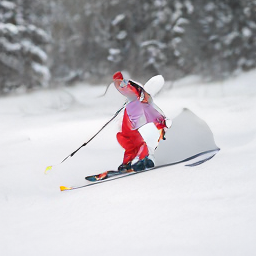} &
        \adjincludegraphics[clip,width=0.13\textwidth,trim={0 0 0 0}]{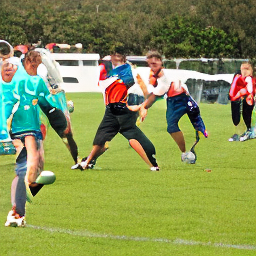} &
        \adjincludegraphics[clip,width=0.13\textwidth,trim={0 0 0 0}]{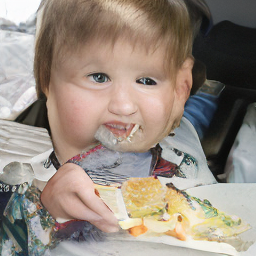} &
        \adjincludegraphics[clip,width=0.13\textwidth,trim={0 0 0 0}]{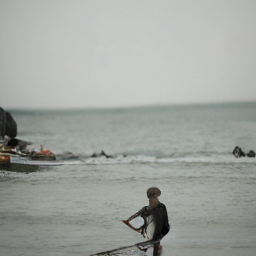}  \\

        \begin{minipage}[t]{0.13\textwidth}
          \tiny ``A man is on a path riding a horse.''
        \end{minipage} &
        \begin{minipage}[t]{0.13\textwidth}
          \tiny ``Tower of a brick building with a clock on the front.''
        \end{minipage} &
        \begin{minipage}[t]{0.13\textwidth}
          \tiny ``A dog leaps through the air as it catches a frisbee.''
        \end{minipage} &
        \begin{minipage}[t]{0.13\textwidth}
          \tiny ``A person in white and red snow suit skiing on a slope.''
        \end{minipage} &
        \begin{minipage}[t]{0.13\textwidth}
          \tiny ``A group of young people playing a game of soccer.''
        \end{minipage} &
        \begin{minipage}[t]{0.13\textwidth}
          \tiny ``A young boy is smiling and he has food around him on a table.''
        \end{minipage} &
        \begin{minipage}[t]{0.13\textwidth}
          \tiny ``a small boy with a hat is standing on a surf board.''
        \end{minipage} \\

    \end{tabular}
    \caption*{\textbf{Figure E: }Qualitative comparison between vanilla, naive curriculum, and ours on MS-COCO dataset.
    }
    \vspace{-6mm}
    \label{fig:coco_t2i}

\end{figure*}
\section{Qualitative Results}

In this section, we present qualitative comparisons between three methods: 1) \textit{Vanilla}, 2) \textit{NaiveCL}, and 3) \textit{Ours}, across the FFHQ, ImageNet, and MS-COCO datasets. All methods are evaluated using DiT-B models, and the final trained models generate all samples shown in the results.
As shown in the results in the following subsections, our approach can synthesize more accurate and realistic images compared to \textit{Vanilla} and \textit{NaiveCL}.

\subsection{Qualitative Evaluation on the FFHQ Dataset.}
Figure \textcolor{red}{C} presents a qualitative analysis of the performance in unconditional facial image synthesis among the vanilla, the naive curriculum approach, and our method. 
Our approach demonstrates superior ability in generating realistic images.
\subsection{Qualitative Analysis on the ImageNet Dataset.}
In the conditional image synthesis, we exhibit the outcomes generated by the vanilla, the naive curriculum strategy, and our proposed method. 
Figure \textcolor{red}{D} clearly shows that our methodology surpasses the competing approaches in terms of quality.
\subsection{Qualitative Assessment on the MS-COCO Dataset.}
To further substantiate the effectiveness of our proposed technique, we conduct a qualitative comparison of the results in the text-to-image generation task among the vanilla, the naive curriculum method, and our own approach, as depicted in Fig. \textcolor{red}{E}.

\section{Further Experimental Results}

\subsection{Convergence Speed across Model Size}

\begin{figure}[h]
    \centering
    \includegraphics[width=0.5\linewidth]{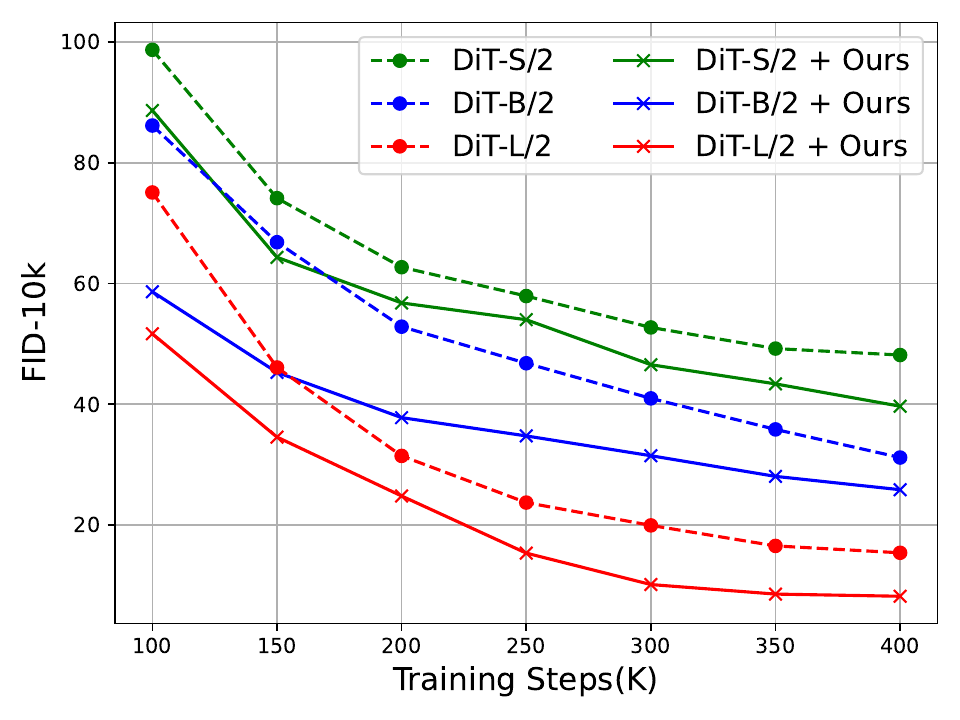}
    \caption*{\textbf{Figure F: }
    We observed an increase in convergence speed across various model sizes when the proposed curriculum learning approach was applied. 
    }
    \label{fig:convergence_appendix}
\end{figure}


By leveraging the advantages of curriculum learning in diffusion training, our method offers faster convergence than vanilla training.
To further investigate this aspect, we measured FID-10K through training iterations for DiT-S and DiT-L.
Figure \textcolor{red}{F} describes the results, showing that our curriculum approach achieves faster convergence in both models.
These results also support the effectiveness of our method.

\subsection{Robustness on Noise Schedule}

\begin{table}[h]
    \centering
     \caption*{\textbf{Table A: }Ablation study on noise scheduler. Note that our approach improves performance consistently across each scheduler.
    }

    \resizebox{0.43\textwidth}{!}{%

    \begin{tabular}{llcccc}
    \toprule
    \multicolumn{6}{l}{\bf{Class-Conditional ImageNet} 256$\times$256.} \\
    \toprule
    Schedule & Method & FID$\downarrow$ & IS$\uparrow$ & Prec$\uparrow$ & Rec$\uparrow$  \\
    \midrule
    \multirow{2}{*}{cosine} & Vanilla & 30.27 & 60.06 & 0.55 & \textbf{0.52} \\
    & Ours & \cellcolor{gray!15}\textbf{22.22} & \cellcolor{gray!15}\textbf{75.98} & \cellcolor{gray!15}\textbf{0.62} &\cellcolor{gray!15}\textbf{0.52} \\
    \cmidrule(lr){1-6}
    \multirow{2}{*}{linear} & Vanilla & 16.99 & 83.62 & 0.68 & \textbf{0.53} \\
    & Ours & \cellcolor{gray!15}\textbf{16.03} & \cellcolor{gray!15}\textbf{87.66} & \cellcolor{gray!15}\textbf{0.69} & \cellcolor{gray!15}\textbf{0.53} \\    
    \bottomrule
    \end{tabular}
    }
   
    \label{tab:schedule}
\end{table}


For a more comprehensive ablation study, we also trained the diffusion model with different noise schedules.
In contrast to cosine scheduling, the $\beta_t$ is set by uniformly dividing the interval $[0.0001, 0.02]$, and the $C_i$ are obtained corresponding to SNR on a linear schedule.
As shown in Table. \textcolor{red}{A}, our approach improves the performance with cosine and linear noise schedulers.

\subsection{Qualitative Results from DiT-L/2 with 2M iterations}
In Figures \textcolor{red}{G}-\textcolor{red}{K}, we present images generated by DiT-L using our curriculum training method for 2M iterations. The results demonstrate that our method produces highly realistic images.

\begin{figure*}[t]
    \centering
    \includegraphics[width=\linewidth]{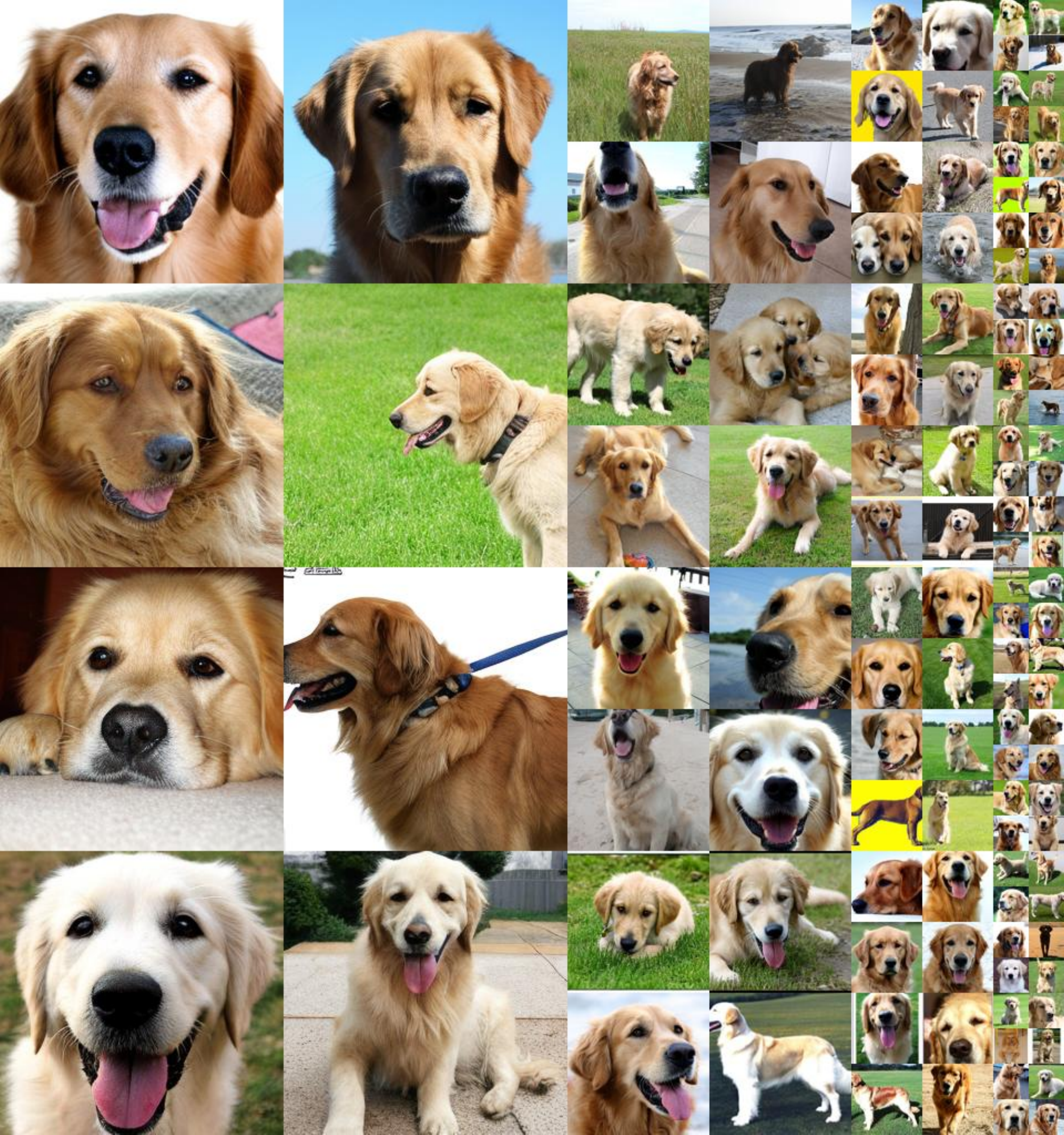}
    \caption*{\textbf{Figure G:} Uncurated 256$\times$256 DiT-L/2 samples. \\ Classifier-free guidanzce scale = 2.0. \\ Class label = ``golden retriever" (207)}
\end{figure*}
\clearpage
\newpage
\begin{figure*}[t]
    \centering
    \includegraphics[width=\linewidth]{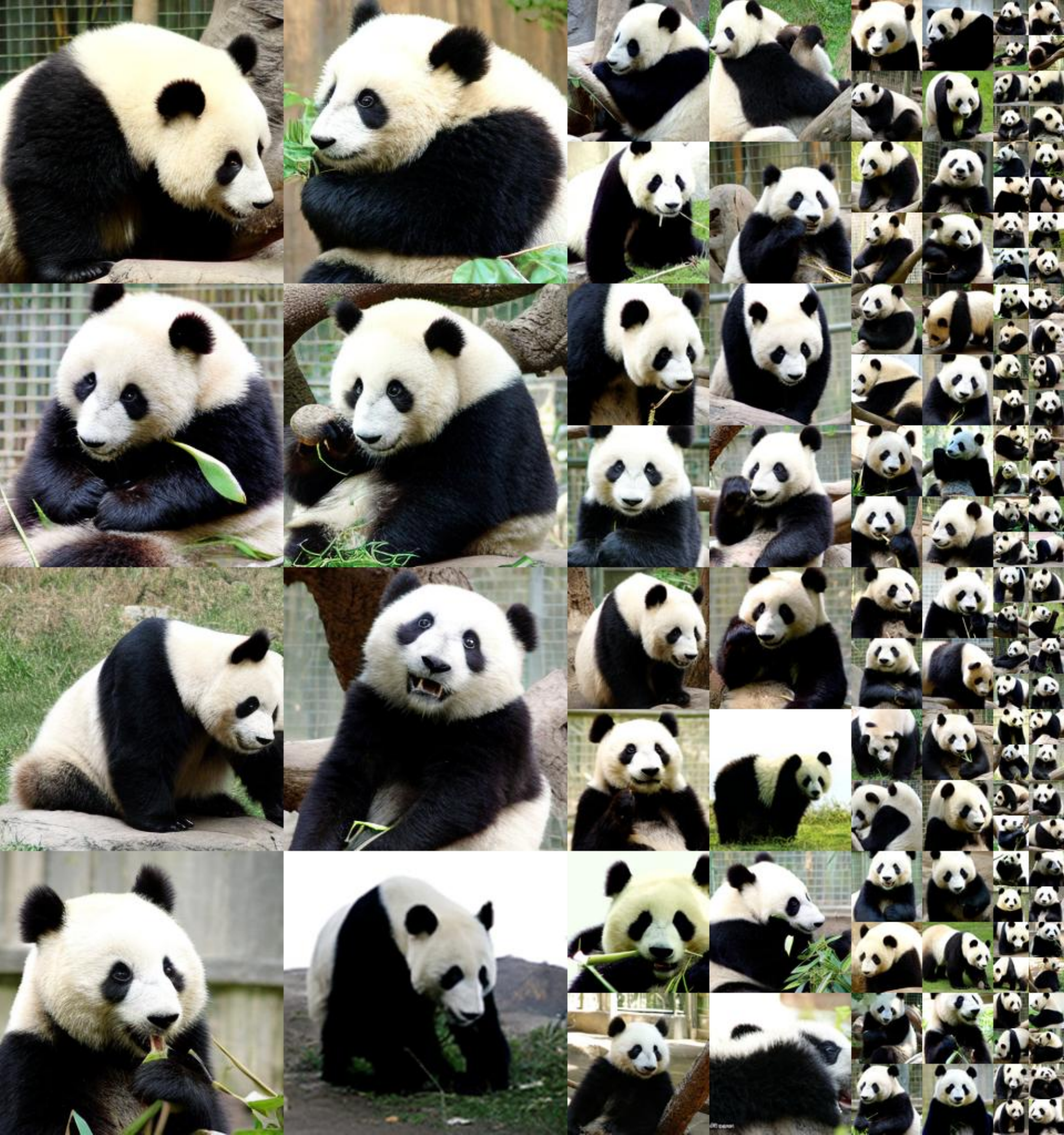}
    \caption*{\textbf{Figure H:} Uncurated 256$\times$256 DiT-L/2 samples. \\ Classifier-free guidance scale = 2.0. \\ Class label = ``panda" (388)}
\end{figure*}
\clearpage
\newpage
\begin{figure*}[t]
    \centering
    \includegraphics[width=\linewidth]{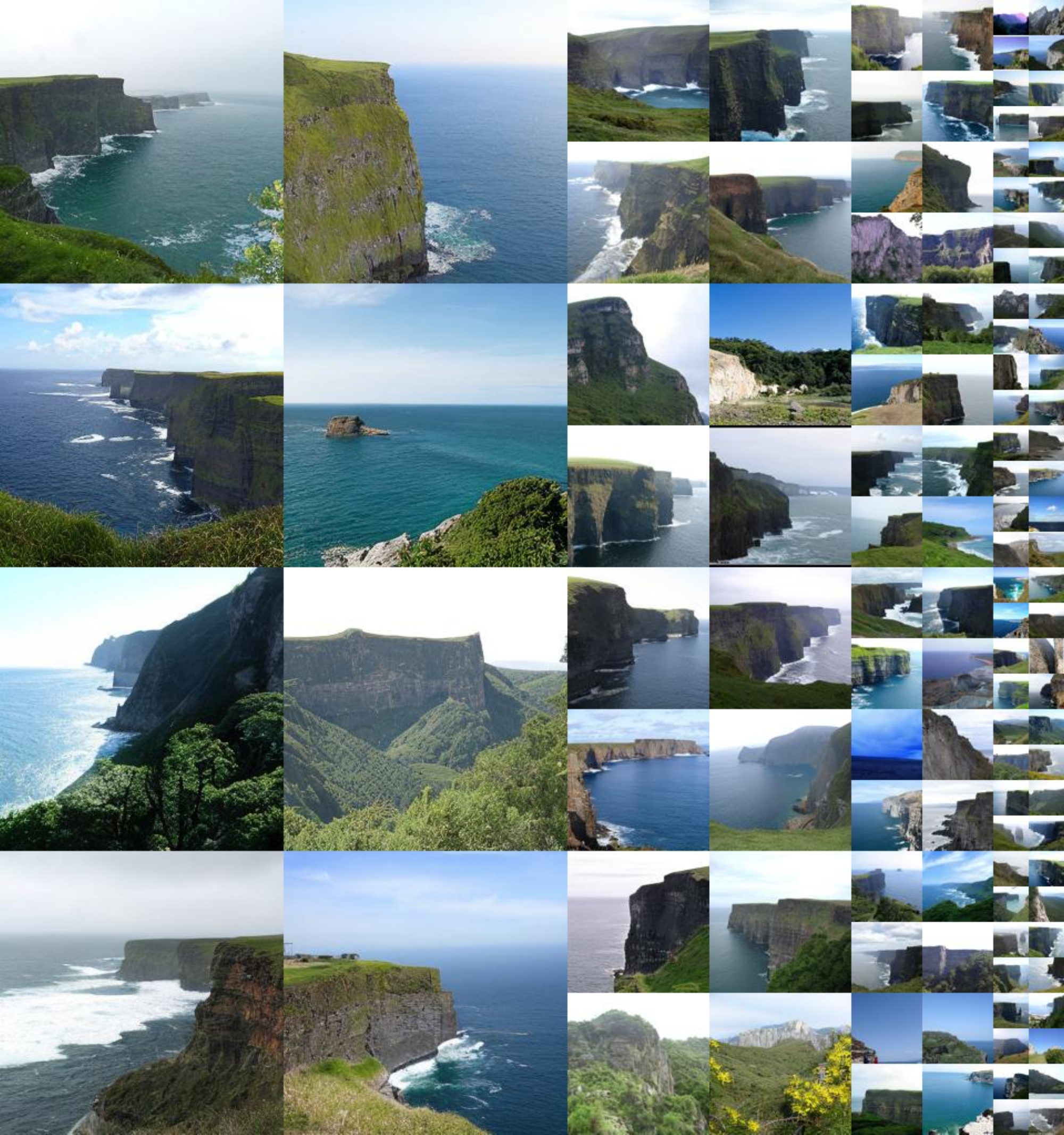}
    \caption*{\textbf{Figure I:} Uncurated 256$\times$256 DiT-L/2 samples. \\ Classifier-free guidance scale = 2.0. \\ Class label = ``cliff drop-off" (972)}
\end{figure*}

\clearpage
\newpage
\begin{figure*}[t]
    \centering
    \includegraphics[width=\linewidth]{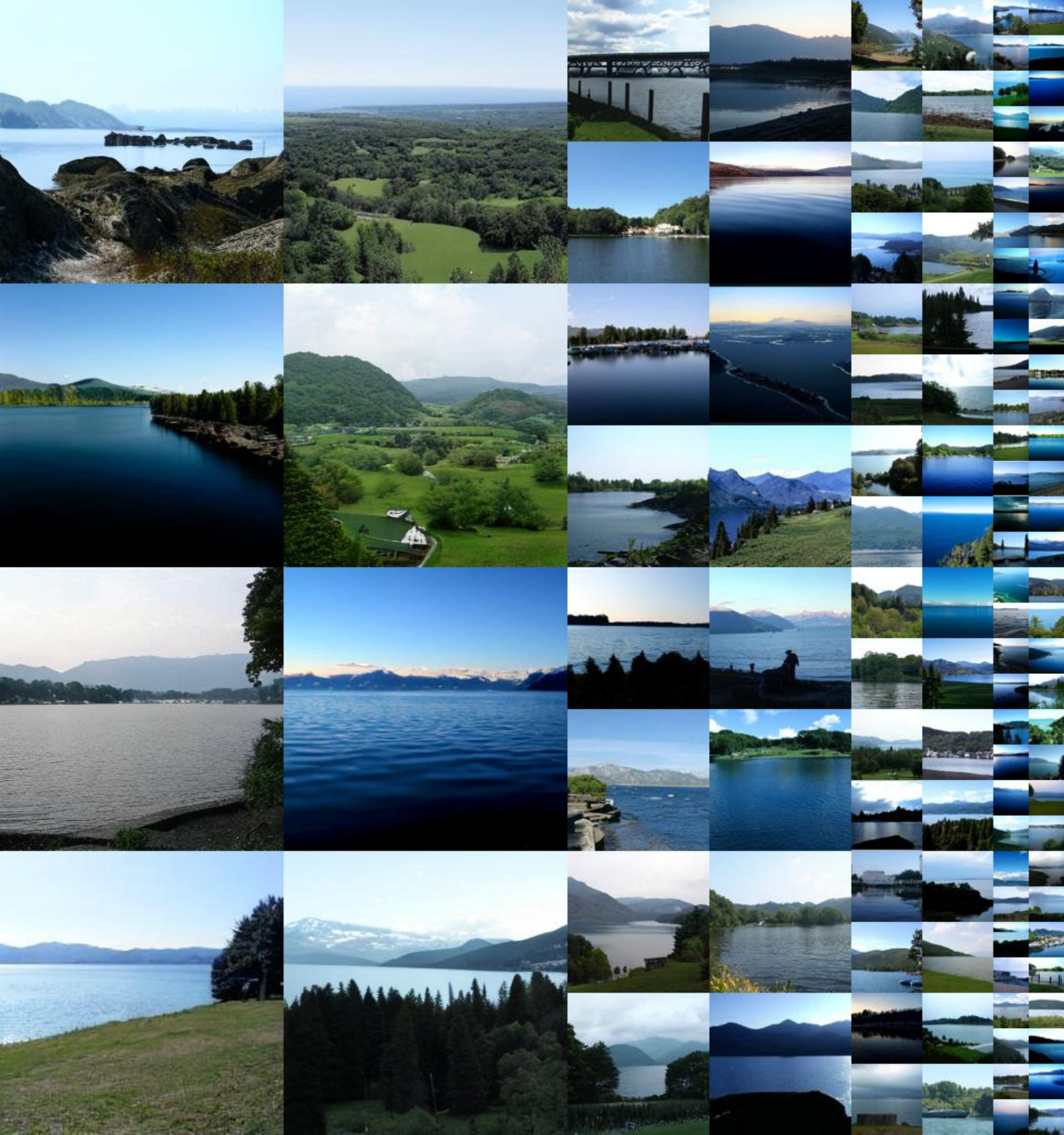}
    \caption*{\textbf{Figure J:} Uncurated 256$\times$256 DiT-L/2 samples. \\ Classifier-free guidance scale = 2.0. \\ Class label = ``lake shore" (975)}
\end{figure*}
\clearpage
\newpage
\begin{figure*}[t]
    \centering
    \includegraphics[width=\linewidth]{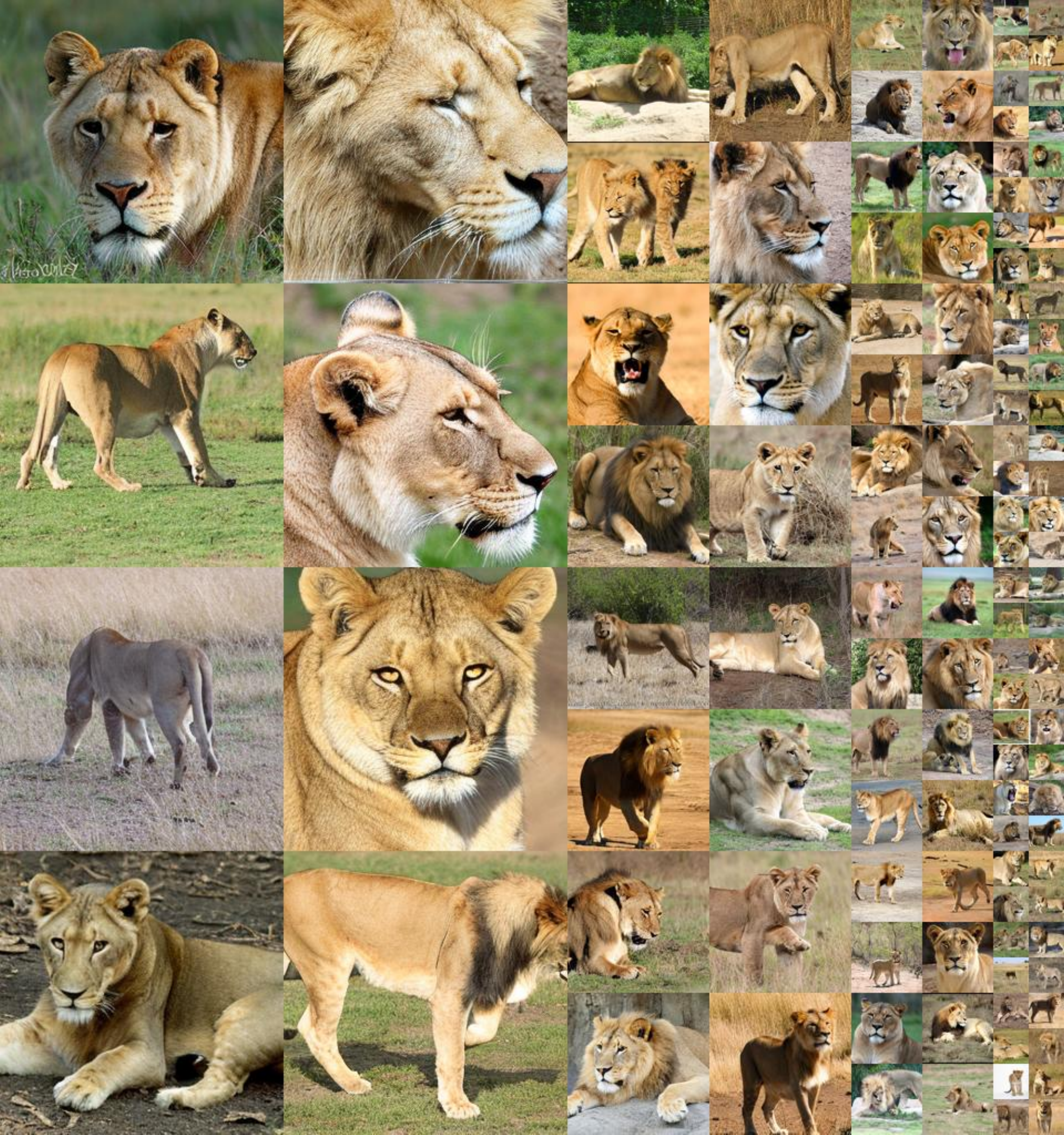}
    \caption*{\textbf{Figure K:} Uncurated 256$\times$256 DiT-L/2 samples. \\ Classifier-free guidance scale = 2.0. \\ Class label = ``lion" (291)}
\end{figure*}

\clearpage
\newpage

\section{Discussion on Similarity with the Previous Work}
While both our work and~\citep{go2023addressing} explore the characteristics of denoising tasks in diffusion models, the aspects of exploration in each work are substantially different.
The notion of task affinity introduced in~\citep{fifty2021efficiently,go2023addressing} refers to how harmoniously the model can learn multiple tasks together. Specifically, their work focuses on identifying and mitigating conflicts between tasks, emphasizing task interactions and transferability by analyzing task similarities (e.g., gradient similarity or alignment).
In contrast, our work explicitly quantifies the relative difficulty of individual denoising tasks across timesteps as a standalone property, independent of task interdependencies. The analysis of task difficulty in our work involves evaluating metrics such as loss behavior or convergence rates, directly reflecting the complexity of solving each task at different timesteps.
Therefore, while~\citep{go2023addressing} addresses how tasks relate and interact during multi-task learning, our focus lies in systematically characterizing the intrinsic difficulty of tasks across timesteps in diffusion models.

\section{Broader Impacts}
Generative models, such as diffusion models, have the potential to significantly impact society, particularly through DeepFake applications and the use of biased datasets. 
One primary concern is the possibility for these models to amplify misinformation, which can erode trust in visual media. 
Additionally, if these models are trained on biased or deliberately altered content, they may unintentionally perpetuate and intensify existing social biases. 
This situation may result in the dissemination of incorrect information and the manipulation of public opinion.

\section{Limitations}

In this work, we demonstrated the varying difficulties of denoising tasks through empirical results on various diffusion frameworks and proposed a curriculum learning approach that effectively enhances diffusion model training. 
While we have shown the robustness of our method's hyperparameters in improving vanilla diffusion training, there is potential for further improvement.
Specifically, curriculum learning methods that utilize smaller hyperparameters and adjust dynamically based on the model itself could yield better results.
We acknowledge the validity of this direction and consider it a promising avenue for future work.

\end{document}